\documentclass[journal,11pt]{IEEEtran}

\usepackage{cite,footnote,xspace,syntonly}
\usepackage{amsmath,mathtools}

\usepackage{cite}
\usepackage{fixltx2e}

\usepackage{graphicx}

\usepackage[utf8]{inputenc}

\usepackage{amsfonts}
\usepackage{amsmath}
\usepackage{mathtools}
\usepackage{amssymb}

\usepackage{bm}

\usepackage{color,verbatim}
\usepackage{multirow}
\usepackage{accents}

\usepackage{theoremref}
\usepackage{flushend}

\usepackage{url}

\newcounter{rulecounter}
\newcommand{\resetrule}{ \setcounter{rulecounter}{0}}
\resetrule

\newsavebox{\selvestebox}
\newenvironment{colbox}[1]
  {\newcommand\colboxcolor{#1}%
   \begin{lrbox}{\selvestebox}%
   \begin{minipage}{\dimexpr\columnwidth-2\fboxsep\relax}}
  {\end{minipage}\end{lrbox}%
   \begin{center}
   \colorbox{\colboxcolor}{\usebox{\selvestebox}}
   \end{center}}

\definecolor{orange}{rgb}{1,0.8,0}
\definecolor{gray}{rgb}{.9,0.9,0.9}
\definecolor{darkgray}{rgb}{.3,0.3,0.3}
\definecolor{darkblue}{rgb}{.1,0.0,0.3}
\definecolor{lightblue}{rgb}{0.7,0.7,1}
\definecolor{lightred}{rgb}{1,0.7,.7}
\definecolor{purple}{RGB}{204,153,255}
\definecolor{lightgray}{rgb}{.95,0.95,0.95}
\definecolor{lightgreen}{rgb}{0.3,0.5,0.3}
\definecolor{darkgreen}{rgb}{0.05,0.3,0.05}

\newcommand{\ra}{$\rightarrow$~}

\newcommand{\hbm}[1]{{\hat{\bm #1}}}
\newcommand{\inv}{^{-1}}
\newcommand{\pinv}{^{\dagger}}

\newcommand{\rfield}{\mathbb{R}}

\newcommand{\diag}[1]{\mathop{\rm diag}\brackets{#1}}

\newcommand{\diagnb}{\mathop{\rm diag}}

 \newcommand{\define}{\triangleq}
\newcommand{\expected}[1]{\mathop{\textrm{E}}\brackets{#1} }
\newcommand{\expectednb}{\mathop{\textrm{E}}\nolimits}

\newcommand{\minimize}{\mathop{\text{minimize}}}

\newtheorem{myproposition}{Proposition}
\newtheorem{myremark}{Remark}
\newtheorem{myproblemstatement}{Problem Statement}
\newtheorem{mylemma}{Lemma}
\newtheorem{mytheorem}{Theorem}
\newtheorem{mydefinition}{Definition}
\newtheorem{mycorollary}{Corollary}

\newcommand{\routingmat}{\hc{\bm \Pi}}
\newcommand{\routingmatentry}{\hc{\Pi}}
\newcommand{\laplacianevalvecspatio}{\laplacianevalvec^{(\spatioind)}}
\newcommand{\laplacianevalvecspatiotemp}{\laplacianevalvec^{(\spatiotempind)}}

\newcommand{\truesignalcov}{\hc{\bm C}}

\newcommand{\adjtransgraphmat}{\hc{\bm A}}
\newcommand{\kernelmatdict}{\hc{\mathcal{D}}}
\newcommand{\timeperiodch}{\hc{t_c}}
\newcommand{\timeperioddel}{\hc{t_d}}
\newcommand{\adjentrynoise}{\hc{\xi}}
\newcommand{\adjentrynoisevar}{\hc{\sigma_A}}

\newcommand{\timetopologychange}{\hc{t_c}}
\newcommand{\seedzeromat}{\hc{\bm D_0}}

\newcommand{\seedmat}{\hc{\bm D}}
\newcommand{\seedmatentry}{\hc{ D}}
\newcommand{\forgetingfactorsynth}{\hc{\delta}}

\newcommand{\fullkernelevaltrvec}{\hc{\tilde{\bm \lambda}}}

\newcommand{\forgetingfactorspatio}{\hc{\gamma}_\spatioind}
\newcommand{\forgetingfactorstatenoise}{\hc{\gamma}_\spatiotempind}
\newcommand{\gradtrvec}{\hc{\bm v}}

\newcommand{\fullkernelevaltr}{\hc{\tilde{\lambda}}}

\newcommand{\spatiocormat}{\hc{\tilde{\bm R}}^{(\spatioind)}}
\newcommand{\spatiocorsmat}{\hc{\bm R}^{(\spatioind)}}

\newcommand{\statenoisecormat}{\hc{\tilde{\bm R}}^{(\spatiotempind)}}
\newcommand{\statenoisecorsmat}{\hc{\bm R}^{(\spatiotempind)}}
\newcommand{\kernelcoefnot}[1]{\hc{(} {#1}\hc{)}}
\newcommand{\kernelcoefspatioreg}{{\hc{\rho}_\spatioind}}
\newcommand{\kernelcoefstatenoisereg}{{\hc{\rho}_\spatiotempind}}
\newcommand{\kernelcoefreg}{\hc{\rho}}
\newcommand{\kernelcoefspatiovec}{\hc{\bm \theta}^{(\spatioind)}}
\newcommand{\kernelcoefspatio}{\hc{\theta}^{(\spatioind)}}
\newcommand{\kernelcoefspatiovecest}{\hc{\hat{\bm \theta}}^{(\spatioind)}}
\newcommand{\kernelcoefstatenoisevec}{\hc{\bm \theta}^{(\spatiotempind)}}
\newcommand{\kernelcoefstatenoise}{\hc{\theta}^{(\spatiotempind)}}
\newcommand{\kernelcoefstatenoisevecest}{\hc{\hat{\bm \theta}}^{(\spatiotempind)}}
\newcommand{\fullkernelspatiomatdict}{\hc{\mathcal{D}}^{(\spatioind)}}

\newcommand{\fullkernelstatenoisematdict}{\hc{\mathcal{D}}^{(\spatiotempind)}}
\newcommand{\omk}{\text{OKM}}
\newcommand{\kkrkf}{\text{KeKriKF}}
\newcommand{\optobjfun}{\hc{\phi}}
\newcommand{\stepsize}{\hc{s}}

\newcommand{\optobjno}[1]{\hc{(}{#1}\hc{)}}
\newcommand{\posprojnot}[1]{\hc{\big[}{#1}\hc{\big]^{+}}}
\newcommand{\itergdnot}[1]{^{#1}}
\newcommand{\itergd}{\hc{k}}
\newcommand{\maxiternumgd}{K}

\newcommand{\genkernelmat}{\hc{\check{\fullkernelmat}}^{(\spatioind)}}

\newcommand{\transweight}{\hc{c}}

\newcommand{\truesignal}{\hc{\bm f}}
\newcommand{\spatiotempind}{\chi}
\newcommand{\spatioind}{\nu}

\newcommand{\truesignalspatiotempcomp}{\hc{\bm f}^{(\spatiotempind)}}
\newcommand{\truesignalspatiotempcompfun}{\hc{ f}^{(\spatiotempind)}}

\newcommand{\truesignalspatiocompfun}{\hc{ f}^{(\spatioind)}}
\newcommand{\truesignalspatiocomp}{\hc{\bm f}^{(\spatioind)}}

\newcommand{\estsignalspatiotempcomp}{\hc{ \hat{\bm f}}^{(\spatiotempind)}}
\newcommand{\estsignalspatiocomp}{\hc{ \hat{\bm f}}^{(\spatioind)}}

\newcommand{\corrmat}{\hc{\bm R}}
\newcommand{\adjacencymatentry}{\hc{A} }
\newcommand{\corrmatprojected}{\hc{\bm T}}
\newcommand{\corrmatprojectedspatio}{\hc{\bm T}^{(\spatioind)}}
\newcommand{\corrmatprojectedspatiotemp}{\hc{\bm T}^{(\spatiotempind)}}

\newcommand{\fullkernelspatiomat}{\hc{\fullkernelmat}^{(\spatioind)}}
\newcommand{\reducedfullkernelspatiomat}{\hc{\bar{\fullkernelmat}}^{(\spatioind)}}
\newcommand{\fullkernelstatenoisemat}{\hc{\fullkernelmat}^{(\spatiotempind)}}

\newcommand{\residualstateervec}{\hc{ \tilde{\bm f}}^{(\spatiotempind)}}

\newcommand{\dlsrstepsize}{\hc{\mu_\text{DLSR}}}
\newcommand{\lmsstepsize}{\hc{\mu_\text{LMS}}}
\newcommand{\dlsrbeta}{\hc{\beta_\text{DLSR}}}
\newcommand{\bandwidth}{\hc{B}}

\newcommand{\kernelindnot}[1]{{{\hc{[}{#1}\hc{]}}}}  % 

\newcounter{exampleind}

\newcounter{remarkind}

\renewcommand{\define}{:=}

\DeclareMathOperator*{\argmin}{arg\,min}

\renewcommand{\expected}[1]{\expectednb\left[#1\right]}  % iteration index 
\renewcommand{\expectednb}{\hc{\mathbb{E}}}

\newcommand{\pdset}{\hc{\mathbb{S}}_+}

\newcommand{\timenot}[1]{_{\hc{}{#1}\hc{}}}  % 
\newcommand{\timegiventimenot}[2]{_{\hc{}{#1}|{#2}\hc{}}}  % 
\newcommand{\timetimenot}[2]{_{\hc{(}{#1},{#2}\hc{)}}}  % 
\newcommand{\timeind}{{\hc{t}}} % time index
\newcommand{\timeset}{\hc{\mathcal{T}}} % time index set
\newcommand{\vertexnot}[1]{_{#1}}  % 
\newcommand{\timevertexnot}[2]{\hc{(}v_{#2},{#1}\hc{)}}  %
\newcommand{\vertexvertexnot}[2]{_{#1,#2}}  %
\newcommand{\timevertexvertexnot}[3]{_{#2,#3}{\hc{(}{#1}\hc{)}}}  %
\newcommand{\graph}{\hc{\mathcal{G}}}

\newcommand{\vertexset}{\hc{\mathcal{V}}}

\newcommand{\edgeset}{\hc{\mathcal{E}}}

\newcommand{\vertexind}{{\hc{{n}}}}

\newcommand{\vertexindp}{{\hc{{\vertexind}'}}} % vertex index prime
\newcommand{\vertexnum}{{\hc{{N}}}}

\newcommand{\adjacencymat}{\hc{\bm A}}

\newcommand{\laplacianmat}{\hc{\bm L}}

\newcommand{\laplacianevecmat}{\hc{\bm U}} 
\newcommand{\laplacianeval}{\hc{\lambda}} 
 
\newcommand{\laplacianevec}{\hc{\bm u}} 
\newcommand{\laplacianevalvec}{\hc{\bm \lambda}}

\newcommand{\signalfun}{{\hc{f}}}

\newcommand{\signalvec}{\hc{\bm \signalfun}}

\newcommand{\fouriersignalfun}{\hc{\fourier f}}

\newcommand{\signalestvec}{\hc{\hbm \signalfun}}

\newcommand{\sampleset}{\hc{\mathcal{S}}} 
\newcommand{\samplemat}{\hc{\bm S}} 
 
\newcommand{\sampleind}{{\hc{s}}} 
\newcommand{\samplenum}{{\hc{S}}} 
 
\newcommand{\extendedsamplenum}{{\hc{\tilde S}}} 
\newcommand{\observationfun}{{\hc{y}}}

\newcommand{\observationvec}{\hc{\bm y}}

\newcommand{\observationnoisefun}{{\hc{e}}} 
\newcommand{\observationnoisevec}{{\bm\observationnoisefun}}

\newcommand{\observationnoisevar}{\hc{ {\sigma^2_\observationnoisefun}}}  % 
\newcommand{\rkhsfunsymbol}{f}
\newcommand{\rkhsvec}{\hc{{\bm \rkhsfunsymbol}}}

\newcommand{\fullkernelmat}{\hc{\bm K}}

\newcommand{\fullkernelevalmat}{\hc{\bm \Lambda_{}}} 
\newcommand{\fullkernelevecmat}{\hc{\bm U_{}}} 
\newcommand{\frequencyweightfun}{\hc{r}}

\newcommand{\rkhsnum}{{\hc{M}}}
\newcommand{\rkhsspationum}{{\hc{M}_\spatioind}}
\newcommand{\rkhsstatenoisenum}{{\hc{M}_\spatiotempind}}
\newcommand{\rkhsind}{{\hc{m}}}
\newcommand{\kernelcoef}{{\hc{ \theta}}}

\newcommand{\kernelcoefvec}{{\hc{\bm \theta}}}
\newcommand{\kernelcoefvecest}{{\hc{\hat{\bm \theta}}}}

\newcommand{\errormat}{{\hc{\bm M}}}
\newcommand{\plantnoise}{\eta} 
 
\newcommand{\plantnoisevec}{\hc{\bm \plantnoise}}

\newcommand{\kalmangainmat}{{\hc{\bm G}}}

\newcommand{\regpar}{\hc{\mu}}

\newcommand{\regparone}{\hc{\mu_1}}
\newcommand{\regpartwo}{\hc{\mu_2}}

\newcommand{\regfun}{\hc{g}}
\newcommand{\graphvariation}{\regfun_\text{LR}}
\newcommand{\laplaciankernelregfun}{\regfun_\text{KR}}

\newcommand{\fourier}[1]{\check{#1}}

\usepackage{epstopdf}
\usepackage{algorithmic}
\usepackage[ruled,vlined,resetcount]{algorithm2e}
\usepackage{subfig}
\usepackage{tabularx}
\usepackage{amsthm}
\usepackage{verbatim}   % for the comment environment
\usepackage{accents}
\usepackage{pgfplots}
\pgfplotsset{compat=newest}
\usetikzlibrary{plotmarks}
\usetikzlibrary{arrows.meta}
\usepgfplotslibrary{patchplots}
\usepackage{grffile}

\pgfplotsset{plot coordinates/math parser=false}
\newlength\mywidth
\newlength\myheight
\definecolor{mycolor1}{rgb}{0.00000,0.44700,0.74100}%
\definecolor{mycolor2}{rgb}{0.85000,0.32500,0.09800}%
\definecolor{mycolor3}{rgb}{0.92900,0.69400,0.12500}%
\definecolor{mycolor4}{rgb}{0.49400,0.18400,0.55600}%
\definecolor{mycolor5}{rgb}{0.46600,0.67400,0.18800}%
\definecolor{mycolor6}{rgb}{0.30100,0.74500,0.93300}%
\definecolor{mycolor7}{rgb}{0.63500,0.07800,0.18400}%

\definecolor{colorKKrKF}{rgb}{0.00000,0.44700,0.74100}%
\definecolor{colorMKrKF}{rgb}{1,0.0032500,0.001}%
\definecolor{colorLMS}{rgb}{0.92900,0.69400,0.12500}%
\definecolor{colorLMS2}{rgb}{0.49400,0.18400,0.55600}%
\definecolor{colorLMS3}{rgb}{0.49400,0.68400,0.55600}%
\definecolor{colorLMS4}{rgb}{0.49400,0.68400,0.95600}%
\definecolor{colorDLSR}{rgb}{0.46600,0.67400,0.18800}%
\definecolor{colorDLSR2}{rgb}{0.30100,0.74500,0.93300}%
\definecolor{colorDLSR3}{rgb}{0.63500,0.07800,0.18400}%

 \newcommand{\cmt}[1]{} % comment
 \newcommand{\hc}[1]{\textcolor{black}{#1}} % highlight command --> to
\newenvironment{myitemize}{}{}
\newcommand{\myitem}{}

\newif\ifshowtikz
\showtikztrue
\showtikzfalse   % <---- comment/uncomment that line

\let\oldtikzpicture\tikzpicture
\let\oldendtikzpicture\endtikzpicture

\renewenvironment{tikzpicture}{%
	\ifshowtikz\expandafter\oldtikzpicture%
	\else\comment%   
	\fi
}{%
\ifshowtikz\oldendtikzpicture%
\else\endcomment%
\fi
}

\theoremstyle{plain}
\newtheorem{proposition}{Proposition}

\newtheorem{mytheoremhere}{Theorem}
\theoremstyle{definition}
\newtheorem{myremarkhere}{Remark}

\newcommand{\transpose}{^{\hc{\top}}}

\newcommand{\identitymat}{\hc{\bm I}}

\DeclareMathOperator{\Tr}{Tr}

\usepackage{fancyhdr}
\begin{document}
\title{Inference of Spatio-Temporal Functions
	over Graphs\\ via Multi-Kernel Kriged Kalman Filtering} %
\author {Vassilis N. Ioannidis$^{\star}$,  \emph{Student Member, IEEE}, Daniel 
Romero$^{\dagger}$, 
\emph{Member, IEEE},\\ and Georgios
  B. Giannakis$^{\star}$, \emph{Fellow, IEEE} \thanks{This work was supported by
   NSF grants 1442686, 1500713, and
    1508993.

$^{\star}$ECE Dept. and the Digital Tech. Center,
 Univ. of Minnesota, Mpls, MN 55455, USA.
 
 $^{\dagger}$ICT Dept., Univ. of Agder, Grimstad 4879, 
 Norway
 
  E-mails: ioann006@umn.edu, daniel.romero@uia.no, georgios@umn.edu}}
\lhead[\fancyplain{\thepage}{\leftmark}]{\tiny{IEEE TRANSACTIONS ON SIGNAL 
		PROCESSING, Nov. 
		2017 (SUBMITTED)}}
\maketitle

\begin{abstract}
	Inference of space-time varying signals on graphs 
	emerges naturally in a plethora of network science 
	related applications. A frequently encountered challenge 
	pertains to reconstructing such dynamic processes, given 
	their values over a subset of vertices and time instants. The 
	present paper develops a graph-aware kernel-based kriged Kalman 
	filter that accounts for the spatio-temporal 
	variations, and offers efficient online reconstruction, even for 
	dynamically evolving network 
	topologies. 
	The  kernel-based learning framework bypasses the need for 
	statistical
	information by capitalizing on the  smoothness that graph signals exhibit
	with respect to the underlying graph. To address the challenge of selecting 
	the appropriate kernel, the 
	proposed filter is combined with a multi-kernel selection module. Such a 
	data-driven method selects a kernel 
	attuned  to 
	the signal 
	dynamics on-the-fly within the linear span of a pre-selected dictionary. 
	The novel multi-kernel learning 
	algorithm exploits
	the eigenstructure of Laplacian kernel matrices to reduce computational 
	complexity.
	Numerical tests with synthetic and real data demonstrate the 
	superior reconstruction performance of the novel approach relative to 
	state-of-the-art 
	alternatives.
\end{abstract}

\section{Introduction}
\label{sec:intro}
\begin{myitemize}
	\myitem\cmt{motivation on graphs-reconstruction} 
	\begin{myitemize}
		\myitem\cmt{graph data}A number of applications involve data that admit a natural representation 
		in terms of node 
		attributes over social, economic, sensor, communication, and biological 
		networks, to name a few~\cite{shuman2013emerging,kolaczyck2009}.  
		\myitem\cmt{reconstruction}An inference task that emerges in this 
		context is to predict or extrapolate 
		the attributes of 
		all nodes in the network given the attributes of a subset of
		them. 
			\myitem\cmt{example}In a finance network, where nodes 
		correspond to 
			stocks 
			and 
			edges capture dependencies among them, one may be interested in 
			predicting 
		the price of all stocks in 
			the 
			network knowing the price of some.
		\myitem\cmt{motivation}This is of paramount importance in 
		applications 	where collecting the attributes of all nodes 
		is prohibitive, as is the case when sampling  large-scale graphs, or, 
		when the 
		attribute of interest is of 
		sensitive nature, such as the transmission of HIV in a social network.  
		\myitem\cmt{graph signal reconstrunction}This task  was first
		formulated as reconstructing a \emph{time-invariant} function on a 
		graph~\cite{shuman2013emerging,smola2003kernels}.
		%, and it is typically 
		%addressed by adopting a time-invariant
		%model that links the topology to the signal values.
%		 Such an inference task
%		typically leverages
%		smoothness of the attributes with respect to the graph, meaning that connected nodes are 
%expected 
%		to have similar 
%		attribute 
%		values.			
	\end{myitemize}
	
	\myitem\cmt{literature review}
	\begin{myitemize}
		\myitem\cmt{reconstruction of time-invariant functions}Follow-up reconstruction approaches 
		leverage the notions of graph bandlimitedness~\cite{anis2016proxies}, sparsity and overcomplete 
		dictionaries~\cite{thanou2014learning}, smoothness over the graph 
		\cite{smola2003kernels,kondor2002diffusion}, all of which can be unified as approximations of 
		nonparametric 
		graph functions drawn from a reproducing kernel Hilbert space 
		(RKHS)~\cite{romero2016multikernel}; see 
		also~\cite{ioannidis2016semipar} for semi-parametric alternatives. 
%		Related  
%		probabilistic  approaches include
%		\emph{network kriging}
%		for predicting Internet path delays 
%		using measurements on a selected subset of paths \cite{chua2006network} .   
		
			\myitem\cmt{motivation on time-varying}In various 
			applications however, 
			the network connectivity and node attributes change over time.\begin{myitemize}
				\myitem\cmt{example}Such is the case in e.g. a
				finance 
				network, where 
				not only the stock prices change over time, but also their inter-dependencies.
				%		 the relations 
				%		among stocks 
				%		as well 
				%		as the stock prices typically vary over time.
				\myitem\cmt{need for tracking dynamics}Hence, maximizing 
				reconstruction performance for these time-varying signals necessitates 
				judicious 
				modeling of the space-time dynamics, especially when samples are 
				scarce. 
			\end{myitemize}

		\myitem\cmt{{reconstruction of time-varying functions}}Inference of 
		\emph{time-varying} graph functions has been so far pursued
		\begin{myitemize}
			\myitem\cmt{no time dynamics}mainly for slow 
			variations~\cite{wang2015distributed,lorenzo2016lms,forero2014dictionary}.
%		\begin{myitemize}\myitem\cmt{wang lorenzo}This is the case with the 
%		least-square method 
%			in~\cite{wang2015distributed}, which requires that the set of sampled nodes be fixed over time. 
%			This limitation is alleviated by the 
%			least mean squares (LMS) estimator 
%			in~\cite{lorenzo2016lms,lorenzo2017lmsrls}, which can track 
%			graph functions from a time-varying and adaptively selected subset of nodes. However, 
%			\cite{wang2015distributed,lorenzo2016lms,lorenzo2017lmsrls} rely on the 
%			bandlimited model, whose effectiveness in capturing the spatial 
%			dynamics of real-world graph functions 
%			is not ensured.
%			\myitem\cmt{forero}This issue is bypassed by the graph-aware dictionary-learning approach
%			in~\cite{forero2014dictionary}, which captures a wide 
%			variety of spatial dynamics but 
%			can only track slow temporal variations.
%				\myitem\cmt{time dynamics}None 
%				of~\cite{forero2014dictionary,wang2015distributed, 
%				lorenzo2016lms,lorenzo2017lmsrls} explicitly
%				 models temporal dynamics and 
%				therefore it is impossible to track rapid changes.
%				%Note that no explicit model for the 
%				%temporal dynamics was utilized 
%				%in~\cite{wang2015distributed,lorenzo2016lms, 
%			%		forero2014dictionary}.
%							\end{myitemize}
							\myitem\cmt{model timedynamics}
							Temporal dynamics have been modeled in \cite{rajawat2014cartography} by assuming 
							that the covariance of the function to be reconstructed is available. On the other hand, 
							spatio-temporal reconstruction of generally dynamic graphs has been approached using 
							an extended graph kernel matrix model with a block tridiagonal structure that lends itself 
							to a computationally 
							tractable iterative 
							solver~\cite{romero2016spacetimekernel}. However, 
							\cite{romero2016spacetimekernel} neither relies on a dynamic model of the function 
							variability, nor it provides a tractable method to learn the ``best" kernel that fits the data.
		\myitem\cmt{limitations}Furthermore, \cite{rajawat2014cartography} and 
		\cite{romero2016spacetimekernel} do not adapt to 
		changes in the 
		spatio-temporal 
		dynamics of the graph function.  
		%To sum up, no low-complexity algorithm that accommodates 
		%general spatio-temporal dynamics and learns them from the data has 
	%	been developed so far.
	\end{myitemize}
		
		%\myitem\cmt{prior work on probabilistic frameworks on graphs}\acom{add?}
		%\begin{myitemize}
		%	\myitem\cite{zhang2015graph} connects Gaussian Markov Random fields with  
		%	undirected graphs \acom{elaborate}.
		%	\myitem\cite{barber2010graphical} puts forth graphical models for time 
		%	series. 
		%	Models such as Kalman filters and hidden Markov models are represented as 
		%	graphical models as well as sequential linear dynamical systems (LDSs).
		%\end{myitemize}
	\end{myitemize}
	
	\myitem\cmt{contributions}The present paper fills this gap by introducing 
	online estimators for time-varying  functions on generally dynamic graphs. Specifically, the 
	contribution is
	threefold.
	\begin{itemize}
		\item[C1.]\cmt{model}A deterministic model for time-varying graph 
		functions is proposed, where 
		spatial dynamics are captured by the network 
				connectivity while temporal dynamics are described through a 
				graph-aware state-space 
				model. 
			
		%capable of capturing a wide variety of spatial dynamics as well as slow 
		%and fast 
		%temporal dynamics. A graph-aware state equation models the temporal evolution of the graph 
		%function.
	 \item[C2.]\cmt{KrKF}Based on this model, an algorithm termed kernel kriged Kalman filter (KeKriKF) 
	 is developed to obtain
	 function estimates by minimizing a kernel ridge 
		regression (KRR) criterion in an online fashion.
		The proposed solver
		generalizes the traditional network kriged Kalman filter
		(KriKF)~\cite{rajawat2014cartography, 
			mardia1998kriged,wikle1999dimension}, which relies on a probabilistic model. 
The novel  
			estimator forgoes with assumptions on data
			distributions and stationarity,
			by promoting space-time smoothness through dynamic  kernels on 
			graphs.

	    	%and 
		%accommodates time-vayring graphs.
		%that encode similarities between the nodes of the graph.
		\item[C3.]\cmt{MKriKF}To select the most appropriate
		kernel, 
		%Multi-kernel learning (MKL) 
		%techniques offer data-driven selection of the pertinent kernel, which 
		%heavily determines the performance of kernel-based 
		%estimators~\cite{romero2016multikernel,romero2016spacetimekernel}. This 
		a 
		multi-kernel (M)KriKF is developed based on the multi-kernel learning (MKL) framework.
			This algorithm adaptively selects the kernel that ``best" fits the 
			data dynamics within the linear span of
			a 
			prespecified kernel
			dictionary.
			%exploiting the available data.
		The structure of Laplacian 
			kernels is exploited to reduce complexity down to the order of 
			KeKriKF. 
		% in the number of nodes in 
		%the network, that renders 
		%the proposed approach attractive 
		%for big data applications. As a convenient byproduct, OKM 
		%The two algorithms KriKF and OKM, are combined to contribute 
		%a novel data-driven approach, termed 
		%multi-kernel KriKF (MKriKF), that performs graph function 
		%reconstruction and MKL in an online 
		%fashion.% Since 
		%effects 
		%The MKriKF in an online fashion and adapts to the observed data 
		%on-the-fly. 
		%an online estimator of graph 
		%functions capable of learning and tracking their 
		%spatio-temporal dynamics from data.To circumvent a novel data-driven approach 
	   %with two capable of selecting the pertinent kernel that 
		%with the challenging problem of multi-kernel learning (MKL). 
%		\myitem\cmt{Laplacian kernel as covariance}A generalization of the latter based on 
%\emph{Laplacian 
%			kernels} is introduced to cope with uncertainty in the spatial statistics of the 
%		process.
		%\myitem\cmt{laplacian kernels GMRF} The novel 
		%\emph{laplacian-kernel}  GMRF (LK-GMRF) is proposed, which can be seen as a 
		%generalization of the intrinsic GMRF (IGMRF)and can be derived from any graph. 
%		\myitem\cmt{Multi-kernel KriKF}Combining the KeKriKF with the MKL algorithm provides a novel 
%		filtering 
%		approach that adapts to the observed data, selects the optimal kernel,  and improves 
%		its future predictions. 
This complexity  is 
		linear in the number of time samples, which renders KeKriKF  and MKriKF appealing for online  
		operation. 
	\end{itemize}

	\myitem\cmt{structure}The rest of the paper is structured as follows. Sec.~\ref{sec:problform} 
	contains  preliminaries and states the problem. 
	Sec.~\ref{sec:mkkrkf} introduces the 
	spatio-temporal model and develops the KeKriKF. Sec.~\ref{sec:omk} endows 
	the KeKriKF  with an MKL module to obtain the  MKriKF. 
	Finally,  
	numerical experiments and conclusions are presented in Secs.~\ref{sec:sims} and~\ref{sec:concl},
	respectively. 
	
	\myitem\cmt{Notation}\emph{Notation:}
	\begin{myitemize}
		\myitem  Scalars are denoted by
		lowercase, column vectors by bold lowercase, and matrices 
		by bold
		uppercase letters. Superscripts $~\transpose$ and $~\pinv$
		respectively 
		denote transpose and pseudo-inverse; $\bm 1_N$ stands for 
		the $N\times1$ all-one vector; $\diag{\bm x}$ corresponds 
		to a diagonal matrix with the entries of $\bm 
		x$ 	on its diagonal, while $\diag{\bm X}$ is a vector holding the diagonal entries of 
		$\bm X$;  and
		$\mathcal{N}(\mu,\sigma^2)$  a Gaussian 
		distribution with mean $\mu$ and variance $\sigma^2$. 
%		  \myitem If $\bm A\define[\bm a_1,\ldots,\bm a_N]$, then $\vect\{\bm
%		  A\}
%		  \define [\bm a_1\transpose,\ldots,\bm a_N\transpose]\transpose\define\bm a$
%		  and  $\vectinv\{\bm a\}\define\bm A$.
%		  \myitem Symbol $\otimes$ denotes the Kronecker product of two matrices 
%		  and the following 
%		  condition holds \acom{$\vect{\{ABC\}}=(C\transpose\otimes A)\vect{\{B\}} $}
		\myitem 
		Finally, if $\bm A$ 
		is a matrix and $\bm x$ a vector, then $\|\bm x\|^2_{\bm
			A}\define \bm x\transpose \bm A\inv \bm x$ and $\|\bm x
		\|_2^2\define \bm x\transpose \bm x$.
		%\acom{define $\ifield^{+}$}
		%			\myitem $(\bm A)_{m,n}$ is the  $(m,n)$-th
		%			entry of matrix $\bm A$.
		%			  \myitem The $n$-th column of the identity matrix
		%			$\identitymat$ is represented by $\canonicalvec{n}$.  \myitem
		%		     \|\bm x\|_{\identitymat}$.  \myitem  $\pdset^N$
		%			represents the cone of $N\times N$ positive definite matrices.
		%			\myitem Finally, $\delta[\cdot]$ stands for the Kronecker delta,
		%			%\myitem $\indicator[C]$ is the indicator of condition $C$, taking 
		%			%the  value 1 if the condition is satisfied and 0 otherwise
		%			\myitem and $\expectednb$ for  expectation.

	\end{myitemize}
	
\end{myitemize}

\section{Problem statement and preliminaries}
\label{sec:problform}
%\subsection{Problem formulation}
\begin{myitemize}
	\myitem\cmt{definitions}
	\begin{myitemize}
		\myitem\cmt{time evolving graph}Consider a time-varying graph
		$\graph\timenot{\timeind}:=(\vertexset,\adjacencymat 
		\timenot{\timeind}),~\timeind=1,2,\ldots$,
		where $\vertexset:=\{v_1, \ldots, v_\vertexnum\}$ denotes the vertex set,
		and $\adjacencymat\timenot{\timeind}$ the
		$\vertexnum\times\vertexnum$ adjacency matrix, whose
		$(\vertexind,\vertexindp)$-th entry 	
		$\adjacencymatentry\timevertexvertexnot{\timeind} 
		{\vertexind}{\vertexindp}$ is the nonnegative 
		weight of the edge connecting vertices $v_\vertexind$ and 
		$v_\vertexindp$ at time $\timeind$.  The edge set is
		$\edgeset\timenot{\timeind} :=\{(v_\vertexind,v_\vertexindp)\in
		\vertexset \times \vertexset:
		\adjacencymatentry\timevertexvertexnot{\timeind} 
		{\vertexind}{\vertexindp}\neq
		0\}$, and two vertices $v$ and $v'$ are 
		\emph{connected} at time $\timeind$ if
		$(v,v')\in \edgeset\timenot{\timeind}$. The graphs $\{\graph\timenot{\timeind}\}_\timeind$ 
		 in this paper are 
		undirected and 
		have 
		no self-loops, which means that	 
		$\adjacencymat\timenot{\timeind}=\adjacencymat\transpose\timenot{\timeind}$ and 
		$\adjacencymatentry\timevertexvertexnot{\timeind}{\vertexind}{\vertexind}=0$, $\forall 
		\timeind,\vertexind$.\begin{myitemize}
			\myitem\cmt{Laplacian}The Laplacian matrix is  $\laplacianmat\timenot{\timeind}:=
			\diag{\adjacencymat\timenot{\timeind}\bm
				1_N}-\adjacencymat\timenot{\timeind}$, and is
				positive 
			semidefinite provided that $\adjacencymatentry\timevertexvertexnot 
			{\timeind}{\vertexind}{\vertexindp}\ge0$, $\forall 
			\vertexind,\vertexindp,\timeind$; see Sec.~\ref{sec:background}.
		\end{myitemize}
		
		\myitem\cmt{time evolving function}A time-varying graph function is a map
		$\signalfun:\vertexset\times\timeset \rightarrow \mathbb{R}$, where
		$\timeset:=\{1,2,\ldots\}$ is the set of time 
		indices. Specifically, $\signalfun\timevertexnot{\timeind}{\vertexind}$ represents the value of the 
		attribute of interest
		at node $\vertexind$ and time 
		$\timeind$, e.g. the closing price of the $\vertexind$-th stock on the $\timeind$-th day. Vector 
		$\truesignal\timenot{\timeind}\define 
		[\signalfun\timevertexnot{\timeind}{1},\ldots,
		\signalfun\timevertexnot{\timeind}{N}]\transpose\in\rfield^\vertexnum$ collects the function 
		values 
		at time $\timeind$.
	\end{myitemize}
%	\acom{rewrite the part without any random assumption}
	
	\myitem\cmt{observation model}Suppose that  ${\samplenum\timenot{\timeind}}$  
	noisy observations 
	$	\observationfun\timevertexnot{\timeind}{\vertexind_\sampleind}= 
	\signalfun\timevertexnot{\timeind}{\vertexind_\sampleind}
	+\observationnoisefun\timevertexnot{\timeind}{\vertexind_\sampleind}
	$, 	$\sampleind=1,\ldots, \samplenum\timenot{\timeind}$, are available at time $\timeind$, 
	%\end{align}
	where $\sampleset\timenot{\timeind}:= 
	\{\vertexind_1,\ldots,\vertexind_{\samplenum\timenot{\timeind}}\}$ contains the indices  
	$1\le\vertexind_1\le\ldots\le\vertexind_{\samplenum\timenot{\timeind}}\le\vertexnum$ of the 
	sampled vertices, and
	$\observationnoisefun\timevertexnot{\timeind}{\vertexind_\sampleind}$ captures the observation 
	error. 
	\begin{myitemize}
		\myitem\cmt{vector model}With $\observationvec\timenot{\timeind}:=[	
		\observationfun\timevertexnot{\timeind}{\vertexind_	1},
			\ldots, 	
			\observationfun\timevertexnot{\timeind} 
			{\vertexind_{\samplenum\timenot{\timeind}}}]$ 
			and $\observationnoisevec\timenot{\timeind}:=[	
			\observationnoisefun\timevertexnot{\timeind}{\vertexind_	1},
			\ldots, 	
			\observationnoisefun\timevertexnot{\timeind} 
			{\vertexind_{\samplenum\timenot{\timeind}}}]$, 
			the observation model in vector-matrix form is
		\begin{align}
		\label{eq:observationsvec}
		\observationvec\timenot{\timeind} 
		=\samplemat\timenot{\timeind}\truesignal\timenot{\timeind} 
		+\observationnoisevec\timenot{\timeind},\quad\timeind=1,2,\ldots
		\end{align}
		where  $\samplemat\timenot{\timeind}\in\{0,1\}^{\samplenum\timenot{\timeind}\times
			\vertexnum}$ selects the sampled entries  of $\truesignal\timenot{\timeind}$.
		%, and 
		%$\observationnoisevec\timenot{\timeind}
		%\in\rfield^{\samplenum\timenot{\timeind}}$ represents the observation 
		%error. 
		
%		It is assumed that $\observationnoisevec\timenot{\timeind}$ has zero mean 
%		$\expectednb[\observationnoisevec\timenot{\timeind} ]=\bm 0$, and is
%		uncorrelated over time and space, meaning that
%		$\expectednb[\observationnoisevec\timenot{\timeind} 
%		\observationnoisevec\transpose\timenot{\tau}]=
%		\observationnoisevar\bm{I}_{\samplenum\timenot{\timeind}}$, if $\timeind=\tau$, and $\bm 
%		0_{\samplenum\timenot{\timeind}, 
%			\samplenum\timenot{\tau}}$ otherwise. 
	
		%    and also a general $\adjtransgraphmat\timetimenot{\timeind}{\timeind-1}$ is 
		%    adopted. 
	\end{myitemize}
	\myitem\cmt{problem statement}Given 
	$\observationvec\timenot{\tau}$, 
	$\samplemat\timenot{\tau}$, and %$\adjtransgraphmat\timetimenot{\tau}{\tau-1}$ and    
	$\adjacencymat\timenot{\tau}$ for 	
	$\tau=1,\ldots,\timeind$, the goal of this paper is to 
	reconstruct 
	$\truesignal\timenot{\timeind}$ at each $\timeind$. The estimators should operate in an online 
	fashion, which means that the computational complexity per time slot 
	$\timeind$ must not grow with $\timeind$.
	%, and still 
	%account 
	%for $\{\observationvec\timenot{\tau},\samplemat\timenot{\tau}, 
	%\adjacencymat\timenot{\tau}\}_{\tau=1}^\timeind$. 
	Observe that no 
	statistical information is assumed available in our formulation. 
	%\myitem\cmt{motivate next}Sec.~\ref{sec:mkkrkf} presents a spatio-temporal 
	%model that captures the 
	%dynamics over space and time and proposes and online and even 
	%data-adaptive solver that  
	%is developed using the kernel-based reconstruction framework; see 
	%Sec.~\ref{sec:background}.
	%perform efficient reconstruction of $\truesignal\timenot{\timeind}$\acom{re}. 
	%	For 
	%	each 
	%	$\observationvec\timenot{\timeind}$ that is observed $\esttruesignal\timenot{\timeind}$ will be 
	%	computed, 	where $\esttruesignal\timenot{\timeind} =\estsignalspatiocomp\timenot{\timeind} 
	%	+\estsignalspatiotempcomp\timenot{\timeind}$, with complexity that does not depend on 
	%	$\timeind$.
\end{myitemize}
%\acom{-------------------------------edit----------------------------------------------}
\subsection{Kernel-based  reconstruction}
\label{sec:background}
\cmt{sec. motivation} 
\cmt{Sec. overview}Aiming ultimately at the time-varying $\signalvec\timenot{\timeind}$, it is 
instructive to outline the kernel-based reconstruction of a time-invariant 
$\signalvec\define[\signalfun\vertexnot{1},\ldots,\signalfun\vertexnot{N}]$ given 
$\graph:=(\vertexset,\adjacencymat)$, and using samples $\observationvec = \samplemat \signalvec
+\observationnoisevec\in\rfield^\samplenum$, where  
$\samplemat\in\{0,1\}^{\samplenum\times\vertexnum}$ and $\samplenum<\vertexnum$.

%\cmt{LS estimator}An immediate approach is to seek a
%least-squares (LS) estimate $\signalestvec
%=\argmin_{\signalvec}||\observationvec-\samplemat\signalvec||_2^2$,
%but since $\samplenum\leq 
%\vertexnum$, the associated system of  linear equations is undetermined 
%and therefore this approach cannot identify $\signalvec$.
%\cmt{reg. estimator}
Relying on  regularized least-squares (LS), we obtain % estimators of the form
\begin{align}\signalestvec
=\argmin_{\signalvec}||\observationvec-\samplemat\signalvec||_2^2+ 
\regpar\regfun(\signalvec)
\end{align}
where $\regpar>0$ and the regularizer $\regfun(\signalvec)$ promotes estimates 
with a
certain structure. 
\cmt{Laplacian reg.}For example,
\begin{myitemize}
	\myitem the so-called \emph{Laplacian} 
	regularizer 
%	\begin{align}
%	\label{eq:graphvariationdef}
	$\graphvariation(\signalvec) \define{(1/2)} \sum_{\vertexind=1}^\vertexnum
	\sum_{\vertexindp=1}^\vertexnum 
	\adjacencymatentry\vertexvertexnot{\vertexind}{\vertexindp} 
	(\signalfun\vertexnot{\vertexind}-\signalfun\vertexnot{\vertexindp})^2$
%	\end{align}
	 promotes smooth function estimates with similar values at
	vertices connected by strong links (large
	$\adjacencymatentry\vertexvertexnot{\vertexind}{\vertexindp}$), since 
    $\graphvariation(\signalvec)$  is small when
		$\signalvec$ is smooth.
%		More 
%		general proxies are reviewed next.
		It turns out that $\graphvariation(\signalvec) =
		\signalvec\transpose\laplacianmat\signalvec$; see
		e.g.~\cite[Ch. 2]{kolaczyck2009}. For a scalar function  $\frequencyweightfun(\laplacianmat)$ a 
		general graph kernel family of regularizers is obtained as $\laplaciankernelregfun(\signalvec) = 
		\signalvec\transpose\fullkernelmat\pinv\signalvec=\|\signalvec 
		\|_{\fullkernelmat}^2$, where %the kernel matrix  and 

	\begin{table}[t]
		\begin{center}
			\begin{tabular}{|p{2.6cm} | p{4cm} | p{1.3cm}|}
				\hline
				\textbf{Kernel name}  & \textbf{Function}  & \textbf{Parameters} \\
				\hline
				\hline
				Diffusion kernel~\cite{kondor2002diffusion}     &
				$r(\lambda)=\exp\{\sigma^2\lambda/2\}$  & 
				$\sigma^2\geq 0$
				\\
				\hline
				$p$-step random walk~\cite{smola2003kernels}    & $r(\lambda) =
				(a-\lambda)^{-p}$ & $a\geq 2$, $p$  \\
				\hline
				Regularized
				Laplacian\cite{smola2003kernels,zhou2004regularization,shuman2013emerging}
				& $r(\lambda)=1 + \sigma^2\lambda$  & $\sigma^2\ge0$ \\
				\hline
				Bandlimited~\cite{romero2016multikernel}    &  
				$\begin{aligned}
				\label{eq:defrbl}
				\frequencyweightfun(\laplacianeval_\vertexind) =  
				\begin{cases}
				1/\beta & 1\le\vertexind\leq\bandwidth\\
				\beta & \text{otherwise}
				\end{cases}
				\end{aligned}$
				&
				$\beta>0$,   $\bandwidth$
				\\
				\hline
					Band-rejection  &  
					$\begin{aligned}
					\label{eq:defrbr}
					\frequencyweightfun(\laplacianeval_\vertexind) =  
					\begin{cases}
					\beta &k\leq \vertexind\leq \vertexnum-l\\
					1/\beta & \text{otherwise}
					\end{cases}
					\end{aligned}$
					&
					$\beta>0$,   $k,$ $l$
					\\
				\hline
			\end{tabular}
		\end{center}
		\caption{Examples of Laplacian kernels and their associated spectral
			weight functions.}
		\label{tab:spectralweightfuns}
	\end{table}
			\begin{figure}[t]
				%\centering{\input{figs/10000.tex}}
				\centering{\includegraphics[width=\linewidth]
					{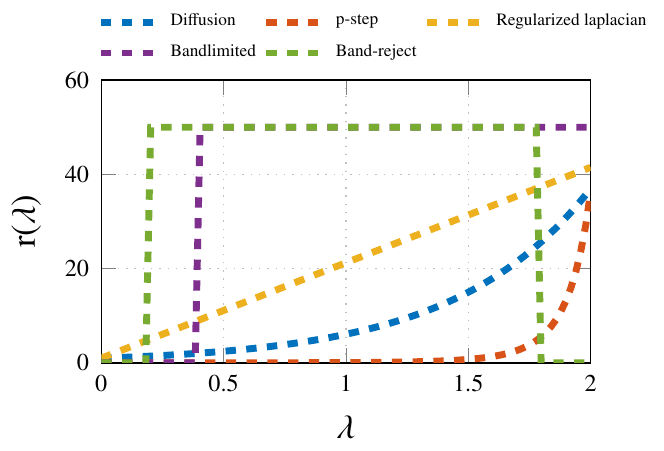}}%\vspace{-1em}
				\hfill
				\caption{Laplacian kernels 
					(Diffusion $\sigma=1.9$, p-step random 
					walk $\alpha=2.55$, $p=6$, 
					Regularized Laplacian $\sigma=4.5$, $\beta=50$, Bandwidth $\bandwidth=20$, $\beta=50$, 
					Band-reject $k=10$, $l=10$).} % temperature estimates. 
				%	($\regparone=1$, 
				%		$\regpartwo=1$, $\dlsrstepsize
				%		=1.6$, $\dlsrbeta=0.5$, $\lmsstepsize =0.6$, 
				%		$\transweight=10^{-3}$, 
				%		$\gausmeanstatenoise=10^{-5}$, $\gausstdstatenoise=10^{-6}$, 
				%		$\gausmeanspatio=2$, 
				%		$\gausstdspatio=0.5$, $\rkhsspationum=40$, $\rkhsstatenoisenum=40$)}
				%\label{fig:recon}
				\label{fig:eigsplot}
				%\vspace{-1em}
			\end{figure}
%			
%		\begin{figure}[t]
%			\centering
%			%	    \hspace{-0.3cm}
%			\includegraphics[width=8.5cm]{eigs.eps}
%			\caption{Laplacian kernels 
%			($\sigma=0.28$ (Diffusion), $\alpha=2.8$, $p=2$ (p-step random 
%			walk), 
%			$\sigma=0.7$ (Regularized Laplacian), $\beta=50$,$\bandwidth=20$ 
%			(Bandlimited), $\beta=50$, $k=10$, $l=10$ (Band-reject).}
%			\label{fig:eigsplot}
%			%\vspace{-2em}
%		\end{figure}
%	
\end{myitemize}

%\cmt{generalization\ra Laplacian kernels}From this perspective, it follows 
%from~\eqref{eq:gvlambda} that $\graphvariation(\signalvec)$ penalizes more 
%heavily
%high-frequency components (large $\laplacianeval_\vertexind$), thus promoting 
%estimates with a
%``low-pass'' graph Fourier transform. A finer control of how
%energy is distributed across 	
%$\{\fouriersignalfun\vertexnot{\vertexind}\}_{\vertexind=1}^\vertexnum$ is 
%available 
%after applying a transformation
%$\frequencyweightfun:\rfield\rightarrow\rfield_+$ to every
%$\laplacianeval\vertexnot{\vertexind}$, giving rise to the \emph{Laplacian 
%kernel} 
%regularizer
%\begin{subequations}
%	\label{eq:generalizedlaplacianregularizer}
%	\begin{align}
%	\label{eq:gvkernel}
%	\laplaciankernelregfun(\signalvec) = \sum_{\vertexind=1}^\vertexnum
%	\frequencyweightfun(\laplacianeval\vertexnot{\vertexind})
%	|\fouriersignalfun\vertexnot{\vertexind}|^2
%	=\signalvec\transpose\fullkernelmat\pinv\signalvec=\|\signalvec 
%	\|_{\fullkernelmat}^2
%	\end{align}
%	where 
	\begin{align}
	\label{eq:laplaciankerneldef}
	\fullkernelmat\define \frequencyweightfun\pinv(\laplacianmat)
	\define\laplacianevecmat\transpose\diagnb\{\frequencyweightfun\pinv(\laplacianevalvec)\}
	\laplacianevecmat
	\end{align}
%\end{subequations}
and is termed a \emph{Laplacian kernel}. Clearly, 
$\laplaciankernelregfun(\signalvec)$ 
subsumes  $\graphvariation(\signalvec)$ for $\frequencyweightfun(\laplacianmat)=\laplacianmat$.
Other special cases of $\laplaciankernelregfun(\signalvec) $ that will be tested in the simulations are 
collected in
Table~\ref{tab:spectralweightfuns}, and the scalar functions are plotted in  Fig~\ref{fig:eigsplot}.  Prior 
knowledge 
about the properties of $\signalvec$
guides the 
selection of the appropriate 
$\frequencyweightfun(\cdot)$, for data-adaptive selection techniques see Sec.~\ref{sec:omk}.

%summarizes prominent examples of kernels on graphs
%arising for specific choices of $\frequencyweightfun$. Prior knowledge 
%about the properties of $\signalvec$, such as its graph Fourier transform 
%$\{\fouriersignalfun\vertexnot{\vertexind}\}_{\vertexind=1}^\vertexnum$, 
%guides the 
%selection of the appropriate 
%$\frequencyweightfun(\cdot)$. For instance, using a diffusion kernel accounts not only for 
%smoothness 
%of $\signalvec$, since it penalizes higher frequency components more heavily 
%(cf.), but also for 
%the prior that $\signalvec$ 
%is generated by 
%a graph diffusion process.
%Similarly, if $\signalvec$ is 
%graph-bandlimited, that is
%$\fouriersignalfun\vertexnot{\vertexind}=0$ for $\vertexind=\bandwidth+1,\ldots,\vertexnum$, one 
%may select 
%the 
%bandlimited kernel, which penalizes the
%frequencies 
%$\vertexind=\bandwidth+1,\ldots,\vertexnum$  (cf. Fig~\ref{fig:eigsplot}).

\cmt{any kernel\ra KRR}Further broadening the scope of the generalized
Laplacian kernel regularizers, one may set
$g(\rkhsvec)=\|\rkhsvec \|_{\fullkernelmat}^2$ for an arbitrary positive 
semidefinite matrix $\fullkernelmat$, not
necessarily a Laplacian kernel. These regularizers give rise to the family of 
\emph{kernel
	ridge regression} (KRR) estimators
\begin{align}
	\signalestvec
:= \argmin_{\rkhsvec} \frac{1}{\samplenum} 
||\observationvec - \samplemat\rkhsvec||^2_2
+ \regpar \|\rkhsvec \|_{\fullkernelmat}^2
\label{eq:krrnotime}
\end{align}
where $\regpar>0$ controls the effect of the regularizer with respect to the
fitting term
${\samplenum}^{-1}||\observationvec-\samplemat\signalvec||_2^2$.
\cmt{advocate KRR}KRR estimators have well-documented merits and solid
grounds on  statistical learning theory; see 
e.g.~\cite{scholkopf2002}. 
%\cmt{more general est.}Different regularizers and 
%fitting functions
%lead to even more general algorithms; see
%e.g.~\cite{romero2016multikernel} for time-invariant graph functions 
%and~\cite{romero2016spacetimekernel} for time-varying graph functions.

\cmt{generalize lmmse}So far, signal $\truesignal$ was assumed deterministic. 
To present a probabilistic interpretation of KRR suppose that 
$\truesignal$ is 
zero-mean with 
$\truesignalcov\define\expected{\truesignal\truesignal\transpose}$, and 
that the entries of $\observationnoisevec$ are uncorrelated with each other and 
with $\signalvec$, and 
$\observationnoisevar\define{\samplenum}\inv 
\expected{\|\observationnoisevec\|_2^2}$. In this setting, 
the 
KRR estimator~\eqref{eq:krrnotime}
reduces to the linear minimum mean-square error (LMMSE) estimator if 
$\regpar\samplenum=\observationnoisevar$ 
and 
$\fullkernelmat=\truesignalcov$. Thus, KRR generalizes 
LMMSE and can be interpreted as the LMMSE estimator of a random signal 
$\truesignal$ with covariance matrix $\fullkernelmat$; see 
~\cite[Proposition 2]{romero2016multikernel}. 
\section{Kernel Kriged Kalman Filter}\label{sec:mkkrkf}
\cmt{Section overview}This section presents a space-time varying model that is capable 
of accommodating fairly general forms of spatio-temporal dynamics.
Building on 
this model,
a novel online KRR estimator will be subsequently developed for graph functions over time-varying 
graphs.
%techniques and captures 
%, and 
%contributes
%an efficient solver for the subproblem of \emph{kernel matching}.
%, that exploits
%the common eigenspace of  Laplacian kernels.

\subsection{Spatio-temporal model}\label{sec:spatiotemp}
An immediate approach to reconstructing $\signalvec\timenot{\timeind}$ 
%in 
%Sec.~\ref{sec:problform} 
is to apply~\eqref{eq:krrnotime}  separately 
per slot~$\timeind$. This yields the instantaneous estimator (IE)
\begin{align}
\label{eq:timeagnostic}
\estsignalspatiocomp\timenot{\timeind}
:=& \argmin_{\rkhsvec} \frac{1}{\samplenum\timenot{\timeind}} 
||\observationvec\timenot{\timeind} - 
\samplemat\timenot{\timeind}\rkhsvec||^2_2
+ \regpar\|\rkhsvec\|_{\fullkernelmat\timenot{\timeind}}^2
\end{align}
where $\fullkernelmat\timenot{\timeind}>\bm 0$ is a per-slot preselected kernel 
matrix, and superscript $\nu$ will be explained later.
% $\estsignalspatiocomp\timenot{\timeind}$ denotes the 
%IE of $\signalvec\timenot{\timeind}$. 
Unfortunately, such an approach does not account for the possible dynamics relating 
$\truesignal\timenot{\timeind}$ to $\truesignal\timenot{\timeind-1}$. However, 
leveraging dependencies across slots can benefit the estimator of 
$\truesignal\timenot{\timeind}$ from 
observations $\{\observationvec\timenot{\tau}\}_{\tau\ne\timeind}$.

\begin{myitemize}
	\myitem\cmt{proposed model}To circumvent the aforementioned limitation, 
	consider modeling the function of 
	interest
	as 	\begin{align}
		\label{eq:decompscalar}
		\signalfun(v_\vertexind,\timeind)= 
		\truesignalspatiocompfun(v_\vertexind,\timeind)
		+\truesignalspatiotempcompfun(v_\vertexind,\timeind) 
	\end{align}
	where 
	$\truesignalspatiocompfun$  captures arbitrary (even fast) temporal dynamics across 
	sampling intervals and can 
	be interpreted as an instantaneous component, while 
	$\truesignalspatiotempcompfun$ represents a structured (typically slow)
	 varying component.
%	modeled. 
%	Specifically, $\truesignalspatiocompfun$  captures instantaneous dependence 
%	among 
%	$\{\signalfun\timevertexnot{\timeind}{\vertexind}\}_{ 
%	\vertexind=1}^\vertexnum$ as the estimator in~\eqref{eq:timeagnostic}, and 
%	$\truesignalspatiotempcompfun$ 
%	 captures dependence between 
%	$\{\signalfun\timevertexnot{\timeind}{\vertexind}\}_{ 
%		\vertexind=1}^\vertexnum$ and its time-lagged version  
%		$\{\signalfun\timevertexnot{\timeind-1}{\vertexind}\}_{ 
%		\vertexind=1}^\vertexnum$.
	\begin{myitemize}
			\myitem\cmt{stock price prediction}As an example, consider 
			stock price prediction, where $\truesignalspatiocompfun$ accounts 
			for
			instantaneous changes caused e.g. by political statements or 
			company 
			announcements at $\timeind$ relative to $\timeind-1$,
			while 
			$\truesignalspatiotempcompfun$ captures the steady 
			evolution of the stock 
			market, 
			where stock prices at slot $\timeind$ are closely related to prices 
			of (possibly) other stocks at $\timeind-1$. 
		\myitem\cmt{decomposition}Before delving into how these components are 
		modeled, let %$\vertexnum\times1$ vectors 
		$\truesignalspatiocomp\timenot{\timeind}\define[ 
		\truesignalspatiocompfun(v_1,\timeind),\ldots, 
		\truesignalspatiocompfun(v_\vertexnum,\timeind)]\transpose$ and
		$\truesignalspatiotempcomp\timenot{\timeind}\define[ 
		\truesignalspatiotempcompfun(v_1,\timeind),\ldots, 
		\truesignalspatiotempcompfun(v_\vertexnum,\timeind)]\transpose$, and 
		note that~\eqref{eq:decompscalar} can 
		be cast into vector form as %This 
		%paper will model 
%		$\signalfun(v_\vertexind,\timeind)= 
%		\truesignalspatiocompfun(v_\vertexind,\timeind)$ $ 
%		+\truesignalspatiotempcompfun(v_\vertexind,\timeind)$
		%	$\truesignal\timenot{\timeind} $
		%	as 
		%	the superpositio
		\begin{align}
		\truesignal\timenot{\timeind} 
		=\truesignalspatiocomp\timenot{\timeind}+\truesignalspatiotempcomp 
		\timenot{\timeind}.
		\label{eq:decomp}
		\end{align}
		Vector
		$\truesignalspatiocomp\timenot{\timeind}$ can be smooth over its entries 		
		$(\graph\timenot{\timeind})$,  and 
		captures  instantaneous dependence 
		among  
		$\{\signalfun\timevertexnot{\timeind}{\vertexind}\}_{ 
			\vertexind=1}^\vertexnum$.
		% , e.g. the 
		%propagation delay in router networks; see 
		%Remark~\ref{re:examp}.
		%and is motivated by the application at hand as the 
		%propagation delay in router networks  
		%and can be viewed as an instantaneous 
		%component that does not depend on its past values 
		%$\{\truesignalspatiocomp\timenot{\timeind-1}\}_\timeind$.
		%affect the 
		%evolution of the function in time.
		\myitem\cmt{model for $\truesignalspatiotempcomp\timenot{\timeind}$}On 
		the other hand, $\truesignalspatiotempcomp\timenot{\timeind}$ is smooth 
		not only over $\graph\timenot{\timeind}$  but also over 
		time, and 	 models dependencies between 
			 $\{\signalfun\timevertexnot{\timeind}{\vertexind}\}_{ 
			 	\vertexind=1}^\vertexnum$ and their time-lagged versions  
			 $\{\signalfun\timevertexnot{\timeind-1}{\vertexind}\}_{ 
			 	\vertexind=1}^\vertexnum$,
		%spatio-temporal
		%dynamics, 
		%e.g. the queuing delay in router networks; see 
		%Remark~\ref{re:examp}.
		\begin{myitemize}			
			\myitem\cmt{state-space}
			%A popular approach ~\cite[Ch. 
			%3]{anderson1958} models
The smooth evolution of $\truesignalspatiotempcomp\timenot{\timeind}$ 
			over time slots adheres 
			to the
			state equation
			\begin{align}
			\truesignalspatiotempcomp\timenot{\timeind}=	
			\adjtransgraphmat\timetimenot{\timeind}{\timeind-1}\truesignalspatiotempcomp\timenot{\timeind-1}
			+ \plantnoisevec\timenot{\timeind}, \quad\timeind=1,2,\ldots
			\label{eq:transmod}
			\end{align}
			where $\adjtransgraphmat\timetimenot{\timeind}{\timeind-1}$ 
			%\acom{rephrase}
			is a \emph{graph} transition matrix,
			% (see 
			%Remark~\ref{re:trans}),
			 and 
			$\plantnoisevec\timenot{\timeind}:=
			[\plantnoise\timevertexnot{\timeind}{1},\ldots, 
			\plantnoise\timevertexnot{\timeind}{\vertexnum} 
			]\transpose\in\rfield^{\vertexnum}$  is termed state noise. Vector 
			$\plantnoisevec\timenot{\timeind}$ will be assumed smooth over 
			$\graph\timenot{\timeind}$, meaning
			$\plantnoise\timevertexnot{\timeind}{\vertexind}$ is expected to be 
			similar to 
			$\plantnoise\timevertexnot{\timeind}{\vertexindp}$ if 
			$\adjacencymatentry\timevertexvertexnot{\timeind}{\vertexind}{\vertexindp}\ne0$.
			% and is 
			%considered smooth over $\graph$.
%			 that 
%			can be interpreted %e.g.
%			 as the  
%			$\vertexnum\times\vertexnum$ adjacency of a possibly directed 
%			"transition graph", with 
%			$\truesignalspatiotempcomp\timenot{0}=\bm 0$, and 
%			$\plantnoisevec\timenot{\timeind}\in\rfield^{\vertexnum}$ capturing the state error.
%			Remark~\ref{re:trans} describes possible choices for 
%			$\adjtransgraphmat\timetimenot{\timeind}{\timeind-1}$.  
			The recursion 
			in~\eqref{eq:transmod} is the graph counterpart of a vector 
			autoregressive model (VARM) of order one (see 
			e.g.~\cite{lutkepohl2005new,shen2016nonlineartmi}), and will lead 
			to computationally efficient online KRR estimators of 
			$\signalvec\timenot{\timeind}$ that account for 
			temporal 
			dynamics~\cite{shen2016nonlineartmi}. 
			%The model in~\eqref{eq:transmod} captures the dependence between
			%$\truesignalspatiotempcompfun(v_\vertexind,\timeind)$ and
			%its time lagged versions 
			%$\{\truesignalspatiotempcompfun(v_\vertexind,\timeind-1)\} 
			%_{\vertexind=1}^\vertexnum$.%, next a model with increased 
			%flexibility is advocated. 
		\end{myitemize}				
		%		\eqref{eq:decomp}, \eqref{eq:transmod}, an online estimator is 
		%		derived in 
		%		Sec.~\ref{sec:kkrkf} that obviates the need for assumptions on data 
		%distributions or knowledge of 
		%		second-order statistics. 
		%		%	The next section~\ref{sec:gkkrkf} presents some backround 
		%	\acom{reconstruction is the general problem }
		%	\acom{break the problem into two}
	\end{myitemize}
%		\myitem\cmt{motivate model}Modeling $\truesignal\timenot{\timeind}$ as the 
%		superposition of a term $\truesignalspatiotempcomp \timenot{\timeind}$ capturing
%		the slow dynamics over time with a state space equation, and a term 
%		$\truesignalspatiocomp \timenot{\timeind}$ accounting for fast dynamics is 
%		motivated by the application at 
%		hand~\cite{wikle1999dimension,rajawat2014cartography,}; 
%		see Remark~\ref{re:examp}.

\begin{myitemize}
	\myitem\cmt{remark}
	%\begin{myremarkhere}
	%	\label{re:examp}
		Model~\eqref{eq:decomp} can be thought of as the graph counterpart of 
		the model adopted in~\cite{wikle1999dimension} to derive the kriged 
		Kalman filter. In our context here,
		$\truesignalspatiocomp\timenot{\timeind}$  describes small-scale 
		\emph{spatial fluctuations} within slot $\timeind$, whereas 
		$\truesignalspatiotempcomp\timenot{\timeind}$ captures
		the so-called \emph{trend} across slots. 
		%The decomposition \eqref{eq:decomp}
		%is often dictated by the sampling interval: whereas 
		%$\truesignalspatiotempcomp\timenot{\timeind}$ captures 
		%slow dynamics relative to the sampling 	interval, fast variations 
		%are %modeled with	
		%$\truesignalspatiocomp\timenot{\timeind}$.
		\myitem\cmt{examples}Furthermore, \eqref{eq:decomp}  generalizes the 
		model used in~\cite{rajawat2014cartography}, where $ 
		\adjtransgraphmat\timetimenot{t}{t-1}=\identitymat_\vertexnum$, for
		\begin{myitemize}
			\myitem\cmt{delay cartography}network 
			delay prediction, where $\truesignalspatiocomp\timenot{\timeind}$ represents the 
			propagation, 
			transmission, 
			and processing delays and
			$\truesignalspatiotempcomp\timenot{\timeind}$ 
			 the 	queuing delay at each router.
		\end{myitemize}	
%	\end{myremarkhere}
\end{myitemize}
	\myitem\cmt{remark}
	\begin{myremarkhere}
		\label{re:trans}
		\begin{myitemize}
			\myitem\cmt{Transition matrix}The transition matrix 
			$\adjtransgraphmat\timetimenot{\timeind}{\timeind-1}$  can be interpreted %e.g.
			as the  	$\vertexnum\times\vertexnum$ adjacency of a 
			generally 
			directed 
			``transition graph" that relates 
			$\{\truesignalspatiotempcompfun\timevertexnot{\timeind-1}{\vertexind}\}_{ 
				\vertexind=1}^\vertexnum$  to
			$\{\truesignalspatiotempcompfun\timevertexnot{\timeind}{\vertexind}\}_{ 
				\vertexind=1}^\vertexnum$.
			%, with 
			%$\truesignalspatiotempcomp\timenot{0}=\bm 0$
			%can be selected in accordance with 
			%the prior 
			%information available.
			Simplicity in estimating $\adjtransgraphmat\timetimenot 
			{\timeind}{\timeind-1}$ motivates the graph version of the random 
			walk 
			model, where $\adjtransgraphmat\timetimenot 
			{\timeind}{\timeind-1}=\transweight\bm 
			I_N$ 
			with $\transweight>0$. On the other hand, adherence to the graph, 
			prompts the selection 
			$\adjtransgraphmat\timetimenot{\timeind}{\timeind-1}= 
			\transweight 
			\adjacencymat$, in which 
			case~\eqref{eq:transmod} 
			%				\acom{mention antonio shift matrix and how 2 relates to a diffusion process}
			amounts to a diffusion process on a time-invariant
			$\graph$.
		\end{myitemize}
	\end{myremarkhere}
\end{myitemize}
\subsection{KeKriKF algorithm}\label{sec:kkrkf}
This section develops an online algorithm  to estimate 
$\signalvec\timenot{\timeind}$, given \eqref{eq:observationsvec} and
$\{\observationvec\timenot{\tau},\samplemat\timenot{\tau}, 
\adjacencymat\timenot{\tau}, 
\adjtransgraphmat\timetimenot{\tau}{\tau-1}\}_{\tau=1}^\timeind$
for the spatio-temporal model of $\signalvec\timenot{\timeind}$ 
in \eqref{eq:decomp} and \eqref{eq:transmod}. 
\begin{myitemize}\myitem\cmt{motivate \eqref{eq:rkhsobj}}
%	To this end, note that the proposed spatio-temporal model from 
%	Sec.~\ref{sec:spatiotemp} accounts for 
%	spatio-temporal dependencies through 
%	$\truesignalspatiotempcomp\timenot{\timeind}$  and 
%	$\truesignalspatiocomp\timenot{\timeind}$, whose knowledge would 
%	immediately provide $\signalvec\timenot{\timeind}$.
	%	to
	%$\truesignalspatiotempcomp\timenot{\timeind-1}$ as well 
	%through~\eqref{eq:decomp} 
	%as $	
	%\truesignal\timenot{\timeind}$ to $\truesignal\timenot{\timeind-1}$ and 
	%accounts for 
	%temporal dynamics of the graph function.  
	Unfortunately, $\{\truesignalspatiocomp\timenot{\tau}$  and 
	$\truesignalspatiotempcomp\timenot{\tau}\}$ cannot be obtained by solving the 
	system of 
	equations 
	comprising~\eqref{eq:observationsvec},~\eqref{eq:decomp}, 
	and~\eqref{eq:transmod} over time even if 
	$\observationnoisevec\timenot{\tau}=\bm 0$ and 
	$\plantnoisevec\timenot{\tau}=\bm 0$ $~\forall\tau$; simply because after 
	replacing $\truesignal 
	\timenot{\tau}$ with $\truesignalspatiotempcomp\timenot{\tau}+
	\truesignalspatiocomp\timenot{\tau}~\forall \tau$,  the estimation task
	involves $2\vertexnum\timeind$ unknowns,  namely
	$\{ 
	\truesignalspatiotempcomp\timenot{\tau},
	\truesignalspatiocomp\timenot{\tau}%,\plantnoisevec\timenot{\tau}
	%\truesignalspatiocompfun\timenot{\tau}\in\rkhs_\spatioind\timenot{\tau},\atop
	%\truesignalspatiotempcompfun\timenot{\tau}\in\rkhs_\spatiotempind\timenot{\tau}
	\}_{\tau=1}^\timeind$,  and only $\extendedsamplenum+\vertexnum\timeind$ 
	equations, where 	
	$\extendedsamplenum\define\sum_{\tau=1}^\timeind\samplenum\timenot{\tau}$ 
	and
	$\extendedsamplenum\le\vertexnum\timeind$.
%	\begin{align}
%	%	 \{\estsignalspatiocomp\timenot{\tau},
%	%	 
%%\estsignalspatiotempcomp\timenot{\tau|\timeind}\}_{\tau=1}^\timeind\define 
%	\underset{\{
%		
%\truesignalspatiotempcomp\timenot{\tau},\plantnoisevec\timenot{\tau},\atop
%		\truesignalspatiocomp\timenot{\tau},\truesignal\timenot{\tau}
%		
%%%\truesignalspatiocompfun\timenot{\tau}\in\rkhs_\spatioind\timenot{\tau},\atop
%		
%%%\truesignalspatiotempcompfun\timenot{\tau}\in\rkhs_\spatiotempind\timenot{\tau}
%		\}_{\tau=1}^\timeind}{\argmin}&
%	~\sum_{\tau=1}^{\timeind}\tfrac{1}{\samplenum\timenot 
%		{\tau}}\|\observationvec\timenot{\tau}
%	-
%	
%\samplemat\timenot{\tau}\truesignal\timenot{\tau}%\truesignalspatiotempcomp\timenot{\tau}
%	%					-\samplemat\timenot{\tau} 
%	%					\truesignalspatiocomp\timenot{\tau}	  
%	\|^2 \nonumber\\
%	\subjectto~~~
%	&\plantnoisevec\timenot{\tau}=\truesignalspatiotempcomp\timenot{\tau}-
%	\adjtransgraphmat 
%	\timetimenot{\tau}{\!\tau-1}\truesignalspatiotempcomp 
%	\timenot{\tau-1}\nonumber\\
%	&\truesignal\timenot{\tau}=\truesignalspatiocomp\timenot{\tau} + 
%	\truesignalspatiotempcomp\timenot{\tau}~~\tau=1,\ldots,\timeind. 
%	%			+\regparthree\|\truesignalspatiotempcomp\transpose\timenot{1} 
%	%			\|^2_{\fullkernelspatiotempmat\timenot{1}}
%	\label{eq:rkhsobjinit}		
%	\end{align}
	To obtain a solution to this underdetermined problem, one must exploit the 
	model structure. Extending the KRR estimator 
	in~\eqref{eq:krrnotime} to time-varying functions, suppose we wish to
	\begin{align}
\label{eq:rkhsobj}
%	 \{\estsignalspatiocomp\timenot{\tau},
%	 \estsignalspatiotempcomp\timenot{\tau|\timeind}\}_{\tau=1}^\timeind\define 
\hspace{-0.2cm}\underset{\{
	\truesignalspatiotempcomp\timenot{\tau},
	\truesignalspatiocomp\timenot{\tau}
	%\truesignalspatiocompfun\timenot{\tau}\in\rkhs_\spatioind\timenot{\tau},\atop
	%\truesignalspatiotempcompfun\timenot{\tau}\in\rkhs_\spatiotempind\timenot{\tau}
	\}_{\tau=1}^\timeind}{\minimize}&
~\sum_{\tau=1}^{\timeind}\tfrac{1}{\samplenum\timenot 
	{\tau}}\|\observationvec\timenot{\tau}
-
\samplemat\timenot{\tau}\truesignalspatiotempcomp\timenot{\tau}
-\samplemat\timenot{\tau}\truesignalspatiocomp\timenot{\tau}	  
\|^2 \\ 
&\hspace{-2cm}+\regparone\sum_{\tau=1}^{\timeind}\|\truesignalspatiotempcomp\timenot{\tau}-
~\adjtransgraphmat\timetimenot{\tau}{\tau-1}\truesignalspatiotempcomp\timenot 
{\tau-1}\|^2_{\fullkernelstatenoisemat\timenot{\tau}}
+\regpartwo\sum_{\tau=1}^{\timeind}\|\truesignalspatiocomp\timenot{\tau}\|^2_ 
{\fullkernelspatiomat\timenot{\tau}}.
%			+\regparthree\|\truesignalspatiotempcomp\transpose\timenot{1} 
%			\|^2_{\fullkernelspatiotempmat\timenot{1}}
\nonumber
\end{align}
%\begin{align}
%%	 \{\estsignalspatiocomp\timenot{\tau},
%%	 \estsignalspatiotempcomp\timenot{\tau|\timeind}\}_{\tau=1}^\timeind\define 
%\label{eq:rkhsobjs}
%\underset{\{
%	\truesignalspatiotempcomp\timenot{\tau},
%	\truesignalspatiocomp\timenot{\tau},\plantnoisevec\timenot{\tau}
%	%\truesignalspatiocompfun\timenot{\tau}\in\rkhs_\spatioind\timenot{\tau},\atop
%	%\truesignalspatiotempcompfun\timenot{\tau}\in\rkhs_\spatiotempind\timenot{\tau}
%	\}_{\tau=1}^\timeind}{\minimize}&
%~\sum_{\tau=1}^{\timeind}\tfrac{1}{\samplenum\timenot 
%	{\tau}}\|\observationvec\timenot{\tau}
%-
%\samplemat\timenot{\tau}\truesignalspatiotempcomp\timenot{\tau}
%-\samplemat\timenot{\tau}\truesignalspatiocomp\timenot{\tau}	  
%\|^2 \nonumber\\ 
%+\regparone&\sum_{\tau=1}^{\timeind}\|
%%\truesignalspatiotempcomp\timenot{\tau}-
%%~\adjtransgraphmat\timetimenot{\tau}{\tau-1}\truesignalspatiotempcomp\timenot 
%%{\tau-1}
%\plantnoisevec\timenot{\timeind}
%\|^2_{\fullkernelstatenoisemat\timenot{\tau}}
%+\regpartwo\sum_{\tau=1}^{\timeind}\|\truesignalspatiocomp\timenot{\tau}\|^2_ 
%{\fullkernelspatiomat\timenot{\tau}}\nonumber\\
%%			+\regparthree\|\truesignalspatiotempcomp\transpose\timenot{1} 
%%			\|^2_{\fullkernelspatiotempmat\timenot{1}}
%\subjectto~~~~&\vspace{-0.3cm}\\
%		\plantnoisevec\timenot{\tau}=&	\truesignalspatiotempcomp\timenot{\tau}-
%			\adjtransgraphmat\timetimenot{\tau} 
%			{\tau-1}\truesignalspatiotempcomp 
%			\timenot{\tau-1}, ~\tau=1,\ldots\timeind\nonumber
%\end{align}
\myitem\cmt{regularizers}where   the 
scalars 
$\regparone,\regpartwo\ge0$ control the trade-off between smoothness and data 
fit, while the regularizers $\|
\truesignalspatiotempcomp\timenot{\tau}-
~\adjtransgraphmat\timetimenot{\tau}{\tau-1}\truesignalspatiotempcomp\timenot 
{\tau-1}
\|^2_{\fullkernelstatenoisemat\timenot{\tau}}$ and 
$\|\truesignalspatiocomp\timenot{\tau}\|^2_ 
{\fullkernelspatiomat\timenot{\tau}}$ effect the smoothness of 
$\plantnoisevec\timenot{\tau}$ and $\truesignalspatiocomp\timenot{\tau}$ 
prescribed by the model. Uncorrelated (nonsmooth) perturbations 
$\plantnoisevec\timenot{\tau}$ can still be captured by setting 
$\fullkernelstatenoisemat\timenot{\timeind}=\identitymat_\vertexnum$, which 
is a Laplacian kernel  with
$\frequencyweightfun(\laplacianeval\vertexnot{\vertexind})=1,~\forall\vertexind$.  When available, 
prior information about 
$\{\truesignalspatiocomp\timenot{\tau},\plantnoisevec\timenot{\tau}\}
_{\tau=1}^\timeind$ may steer the selection of suitable kernel matrices; when 
not available, one can resort to the algorithm in Sec.~\ref{sec:omk}.
%				\myitem\cmt{rewrite}The constraint in \eqref{eq:rkhsobjs} is 
%				needed to enforce the state space 
%				model~\eqref{eq:transmod}. This constraint can be removed by 
%				eliminating 
%				$\plantnoisevec\timenot{\tau}$ to arrive at
				
				Directly solving~\eqref{eq:rkhsobj} per $\timeind$ 
				would not lead to an online algorithm since the complexity of 
				such an approach grows  with $\timeind$; see 
				Sec.~\ref{sec:problform}.  
				However, we will develop next an efficient \emph{online} 
				algorithm to obtain per slot $\timeind$ estimates
				$		 
				\estsignalspatiotempcomp\timenot{\timeind|\timeind},
				\estsignalspatiocomp\timenot{\timeind|\timeind}%\}_{\tau=1}^\timeind
				$ that still account for $\{\observationvec\timenot{\tau},\samplemat\timenot{\tau}, 
				\adjacencymat\timenot{\tau}\}_{\tau=1}^\timeind$.
				
							\begin{myitemize}
								\myitem\cmt{differentiate wrt $\truesignalspatiocomp\timenot{\tau}$}
								Given $\truesignalspatiotempcomp\timenot{\tau}$,  the first-order necessary 
								conditions for 
								optimality of 
								$\truesignalspatiocomp\timenot{\tau}$ 
								yield [cf. \eqref{eq:rkhsobj}]
								\begin{align}
								\truesignalspatiocomp\timenot{\tau}=%&
								\fullkernelspatiomat\timenot{\tau} 
								\samplemat\transpose\timenot{\tau} ( \reducedfullkernelspatiomat\timenot{\tau}+ 
								\regpartwo\samplenum\timenot{\tau}\bm 
								I_{\samplenum\timenot{\tau}})\inv
								%\nonumber\\\times&
								(\observationvec\timenot{\tau}- 
								\samplemat\timenot{\tau}\truesignalspatiotempcomp\timenot{\tau})
								\label{eq:defofspat}
								\end{align}
								where 
								$\reducedfullkernelspatiomat\timenot{\tau}:= \samplemat\timenot{\tau} 
								\fullkernelspatiomat\timenot{\tau}  
								\samplemat\transpose\timenot{\tau}$. Notice that the overbar notation 
								indicates 
								$\samplenum\timenot{\tau}\times\samplenum\timenot{\tau}$ 
								matrices or $\samplenum\timenot{\tau}\times1$ vectors, and recall that without 
								overbar their counterparts have sizes $\vertexnum\times\vertexnum$ and 
								$\vertexnum\times1$, respectively.
								\myitem\cmt{Derive equivelant objective without 
								$\truesignalspatiocomp$}Substituting~\eqref{eq:defofspat} into~\eqref{eq:rkhsobj},
								we 
								arrive at an 
								optimization problem that does not depend on $\truesignalspatiocomp\timenot{\tau}$ 
								for 
								$\tau=1, \ldots, \timeind$. Rewrite next the per slot  $\tau$ measurement error 
								in~\eqref{eq:rkhsobj} using~\eqref{eq:defofspat} as
								\begin{subequations}
								\begin{align}
								&\tfrac{1}{\samplenum\timenot 
									{\tau}}\|\observationvec\timenot{\tau}
								-
								\samplemat\timenot{\tau}\truesignalspatiotempcomp\timenot{\tau}
								-\samplemat\timenot{\tau}\truesignalspatiocomp\timenot{\tau}	  
								\|^2 \nonumber\\ 
								=&
								\tfrac{1}{\samplenum\timenot 
									{\tau}}\|
								\observationvec\timenot{\tau}- 
								\samplemat\timenot{\tau}\truesignalspatiotempcomp\timenot{\tau} 
								-\reducedfullkernelspatiomat\timenot{\tau}
								\nonumber\\ 
								\times&
								(\reducedfullkernelspatiomat\timenot{\tau}+\regpartwo\samplenum\timenot{\tau}
								\identitymat_{\samplenum\timenot{\tau}})\inv(\observationvec\timenot{\tau}- 
								\samplemat\timenot{\tau}\truesignalspatiotempcomp\timenot{\tau})\|^2\nonumber\\
								=&\tfrac{1}{\samplenum\timenot 
									{\tau}}\|\big[
								\identitymat_{\samplenum\timenot{\tau}}- 
								\reducedfullkernelspatiomat\timenot{\tau}(\reducedfullkernelspatiomat\timenot{\tau}+
								\regpartwo\samplenum\timenot{\tau}
								\identitymat_{\samplenum\timenot{\tau}})\inv \big]\nonumber\\ 
								\times&
								(\observationvec\timenot{\tau}- 
								\samplemat\timenot{\tau}\truesignalspatiotempcomp\timenot{\tau})\|^2.
								\label{eq:difinter}
								\end{align}
							 The matrix inversion lemma asserts for the matrix in square brackets of \eqref{eq:difinter}
							 	that
							 	\begin{align}
							 	&\big[
							 	\identitymat_{\samplenum\timenot{\tau}}- 
							 	\reducedfullkernelspatiomat\timenot{\tau}(\reducedfullkernelspatiomat\timenot{\tau}+
							 	\regpartwo\samplenum\timenot{\tau}
							 	\identitymat_{\samplenum\timenot{\tau}})\inv \big]
							 	\nonumber\\=&(\identitymat_ 
							 	{\samplenum\timenot{\tau}}+\tfrac{1}{\regpartwo\samplenum\timenot{\tau}}
							 	\reducedfullkernelspatiomat\timenot{\tau})\inv.
							 	\label{eq:mil}
							 	\end{align}
							 	Plugging \eqref{eq:mil} into \eqref{eq:difinter} yields 
								\begin{align}
								=&\tfrac{1}{\samplenum\timenot 
									{\tau}}\|(\tfrac{1}{\regpartwo\samplenum\timenot{\tau}}
								\reducedfullkernelspatiomat\timenot{\tau}+\identitymat_ 
								{\samplenum\timenot{\tau}})\inv
								(\observationvec\timenot{\tau}- 
								\samplemat\timenot{\tau}\truesignalspatiotempcomp\timenot{\tau})\|^2\nonumber\\
								=&(\observationvec\timenot{\tau}- 
								\samplemat\timenot{\tau}\truesignalspatiotempcomp\timenot{\tau})\transpose ( 
								\tfrac{1}{\regpartwo}\reducedfullkernelspatiomat\timenot{\tau}+ 
								\samplenum\timenot{\tau}\bm 
								I_{\samplenum\timenot{\tau}})^{-\top}\nonumber\\
								\times&
								\samplenum\timenot{\tau}\bm 
								I_{\samplenum\timenot{\tau}}( 
								\tfrac{1}{\regpartwo}\reducedfullkernelspatiomat\timenot{\tau}+ 
								\samplenum\timenot{\tau}\bm 
								I_{\samplenum\timenot{\tau}})\inv(\observationvec\timenot{\tau}- 
								\samplemat\timenot{\tau}\truesignalspatiotempcomp\timenot{\tau}).
								\label{eq:difresult}
								\end{align}
							
								Next, we express  the regularizer in~\eqref{eq:rkhsobj}
								using~\eqref{eq:defofspat} for each $\tau$ as 
								\begin{align}
								&\regpartwo\|\truesignalspatiocomp\timenot{\tau}\|^2_ 
								{\fullkernelspatiomat\timenot{\tau}}\nonumber\\
%								=&\regpartwo
%								\|\fullkernelspatiomat\timenot{\tau} 
%								\samplemat\transpose\timenot{\tau} ( \reducedfullkernelspatiomat\timenot{\tau}+ 
%								\regpartwo\samplenum\timenot{\tau}\bm 
%								I_{\samplenum\timenot{\tau}})\inv\nonumber\\\times&
%								(\observationvec\timenot{\tau}- 
%								\samplemat\timenot{\tau}\truesignalspatiotempcomp\timenot{\tau})\|^2_ 
%								{\fullkernelspatiomat\timenot{\tau}}\nonumber\\
%								=&\regpartwo(\observationvec\timenot{\tau}- 
%								\samplemat\timenot{\tau}\truesignalspatiotempcomp\timenot{\tau})\transpose ( 
%								\reducedfullkernelspatiomat\timenot{\tau}+ 
%								\regpartwo\samplenum\timenot{\tau}\bm 
%								I_{\samplenum\timenot{\tau}})^{-\top}\nonumber\\\times&
%								\samplemat\timenot{\tau}{\fullkernelspatiomat\timenot{\tau}}\transpose
%								{\fullkernelspatiomat\timenot{\tau}}\inv
%								\fullkernelspatiomat\timenot{\tau} 
%								\samplemat\transpose\timenot{\tau} ( \reducedfullkernelspatiomat\timenot{\tau}+ 
%								\regpartwo\samplenum\timenot{\tau}\bm 
%								I_{\samplenum\timenot{\tau}})\inv\nonumber\\\times&
%								(\observationvec\timenot{\tau}-
%								 								
%\samplemat\timenot{\tau}\truesignalspatiotempcomp\timenot{\tau})\nonumber\\
								=&(\observationvec\timenot{\tau}- 
								\samplemat\timenot{\tau}\truesignalspatiotempcomp\timenot{\tau})\transpose ( 
								\tfrac{1}{\regpartwo}\reducedfullkernelspatiomat\timenot{\tau}+ 
								\samplenum\timenot{\tau}\bm 
								I_{\samplenum\timenot{\tau}})^{-\top}\nonumber\\\times&
								\tfrac{1}{\regpartwo}\reducedfullkernelspatiomat\timenot{\tau}( 
								\tfrac{1}{\regpartwo}\reducedfullkernelspatiomat\timenot{\tau}+ 
								\samplenum\timenot{\tau}\bm 
								I_{\samplenum\timenot{\tau}})\inv(\observationvec\timenot{\tau}- 
								\samplemat\timenot{\tau}\truesignalspatiotempcomp\timenot{\tau})
								\label{eq:regtransf}
								\end{align}				
								where the last equality follows from
								the definition of $\reducedfullkernelspatiomat\timenot{\tau}$.
				          \end{subequations} Combining 
								\eqref{eq:difresult} with \eqref{eq:regtransf} yields
								\begin{align}
								&\tfrac{1}{\samplenum\timenot 
									{\tau}}\|\observationvec\timenot{\tau}
								-
								\samplemat\timenot{\tau}\truesignalspatiotempcomp\timenot{\tau}
								-\samplemat\timenot{\tau}\truesignalspatiocomp\timenot{\tau}	  
								\|^2+\regpartwo\|\truesignalspatiocomp\timenot{\tau}\|^2_ 
								{\fullkernelspatiomat\timenot{\tau}}\nonumber\\
%								=&(\observationvec\timenot{\tau}- 
%								\samplemat\timenot{\tau}\truesignalspatiotempcomp\timenot{\tau})\transpose ( 
%								\tfrac{1}{\regpartwo}\reducedfullkernelspatiomat\timenot{\tau}+ 
%								\samplenum\timenot{\tau}\bm 
%								I_{\samplenum\timenot{\tau}})^{-\top}\nonumber\\
%								 \times&( 
%								\tfrac{1}{\regpartwo}\reducedfullkernelspatiomat\timenot{\tau}+ 
%								\samplenum\timenot{\tau}\bm 
%								I_{\samplenum\timenot{\tau}})( 
%								\tfrac{1}{\regpartwo}\reducedfullkernelspatiomat\timenot{\tau}+ 
%								\samplenum\timenot{\tau}\bm 
%								I_{\samplenum\timenot{\tau}})\inv\nonumber\\
%								\times&(\observationvec\timenot{\tau}- 
%								\samplemat\timenot{\tau}\truesignalspatiotempcomp\timenot{\tau})\nonumber\\
								=&\|\observationvec\timenot{\tau}- 
								\samplemat\timenot{\tau} 
								\truesignalspatiotempcomp\timenot{\tau}\|^2_{\genkernelmat\timenot{\tau}}
								\label{eq:fittermreform}
								\end{align}
								where 
								$\genkernelmat\timenot{\tau}:=\tfrac{1}{\regpartwo} 
								\reducedfullkernelspatiomat\timenot{\tau}+ 
								\samplenum\timenot{\tau}\identitymat_
								{\samplenum\timenot{\tau}}$. 
								\myitem\cmt{Reformulate~\eqref{eq:rkhsobj}}
								Using~\eqref{eq:fittermreform} per slot, \eqref{eq:rkhsobj} boils down to
								\begin{align}
								\label{eq:kfobj}
								\{\estsignalspatiotempcomp\timenot{\tau|\timeind}\}^\timeind_{\tau=1}\define
								\underset{\{
									\truesignalspatiotempcomp\timenot{\tau}
									%\truesignalspatiotempcompfun\timenot{\tau}\in 
									%\rkhs_\spatiotempind\timenot{\tau}
									\}_{\tau=1}^\timeind}{\argmin}
								& \sum_{\tau=1}^{\timeind}\|\observationvec\timenot{\tau}- 
								\samplemat\timenot{\tau} 
								\truesignalspatiotempcomp\timenot{\tau} \|^2_{\genkernelmat\timenot{\tau}}\\
								+&\regparone \sum_{\tau=1}^{\timeind}\|\truesignalspatiotempcomp\timenot{\tau}-
								~\adjtransgraphmat\timetimenot{\tau}{\tau-1}\truesignalspatiotempcomp\timenot 
								{\tau-1}\|^2_{\fullkernelstatenoisemat\timenot{\tau}}.
								%				+&\regparthree\truesignalspatiotempcomp\transpose\timenot{1} 
								%				\fullkernelspatiotempmat\inv\timenot{1} 
								%				\truesignalspatiotempcomp\timenot{1}
								\nonumber
								\end{align}
								Since \eqref{eq:kfobj} is identical to the deterministic formulation of the Kalman filter 
								(KF) applied to a state-space model with state noise covariance 
								$\fullkernelstatenoisemat\timenot{\timeind}$ and measurement noise covariance 
								$\genkernelmat\timenot{\timeind}$, we deduce that the KF algorithm, see e.g. \cite[Ch. 
								17]{strang1997linear}, applies readily to obtain sequentially the structured per slot 
								$\timeind$ 
								component $\{\estsignalspatiotempcomp\timenot{\tau|\tau}\}_{\tau=1}^\timeind$. 
								After 
								substituting $	
								\{\estsignalspatiotempcomp\timenot{\tau|\tau}\}^\timeind_{\tau=1}$ 
								into~\eqref{eq:defofspat},
								we can find also  the per slot instantaneous component $	
								\{\estsignalspatiocomp\timenot{\tau|\tau}\}^\timeind_{\tau=1}$.
							\myitem\cmt{KeKriKF}The $\timeind$-th iteration of our so-termed KeKriKF is 
							listed as
							Algorithm~\ref{algo:krkalmanfilter}. 
								
								Summing up, we have established the following result.
								%				\acom{Should we have 
								%					$\estsignalspatiocomp\timenot{\tau|\tau}$? Yes change everywhere}
							\end{myitemize}

		\myitem\cmt{Derive KeKriKF from \eqref{eq:rkhsobj} }
		\begin{mytheoremhere}\thlabel{thm:kkrkf}
			If 
			$\big\{\{\estsignalspatiotempcomp\timegiventimenot{\tau}{\timeind}, 
			\estsignalspatiocomp\timegiventimenot{\tau}{\timeind}\}_{\tau=1}^{\tau=\timeind}\big\}
			_{\timeind=1}^{\timeind=\timeind_1}$ solves \eqref{eq:rkhsobj} for 
			$\timeind=1,\ldots,\timeind_1$, the KeKriKF iterations summarized in 
			Algorithm~\ref{algo:krkalmanfilter} for 
			$\timeind=1,\ldots,\timeind_1$
			generate the subset of solutions 
			$\{\estsignalspatiotempcomp\timegiventimenot{\timeind}{\timeind}, 
			\estsignalspatiocomp\timegiventimenot{\timeind}{\timeind}\}_ 
			{\timeind=1}^{\timeind=\timeind_1}$.
		\end{mytheoremhere}
%		\begin{IEEEproof}
%			See Appendix~\ref{proof:thkrkf}.
%		\end{IEEEproof}
%	Its complexity is 
%	$\mathcal{O}(\vertexnum^3)$, which does not depend on $\timeind$;  for large $\vertexnum$ see 
%	Remark~\ref{re:complex}. 
	%Therefore,
	%Algorithm~\ref{algo:krkalmanfilter} is an online algorithm. 
	Clearly, 
	the  KeKriKF algorithm comprises two 
	subprocedures:
	Kalman filtering (steps S1 - S6), % of 
	%Algorithm~\ref{algo:krkalmanfilter}
	  and  kriging (step S7). % of 
	%Algorithm~\ref{algo:krkalmanfilter}; 
	\begin{myitemize}

		\myitem\cmt{differences}
	\begin{myitemize}
		\myitem\cmt{difference 
		with general model 
		for KriKF}The traditional KriKF has been employed to 
	interpolate 
	stationary processes defined over continuous spatial domains 
	\cite{mardia1998kriged,wikle1999dimension}, and its derivation follows from a probabilistic 
	linear-minimum mean-square error (LMMSE) criterion that relies on knowledge 
	of second-order 
	statistics~\cite{rajawat2014cartography, 
		mardia1998kriged,wikle1999dimension}. Here, our KeKriKF
		is derived from a deterministic kernel-based learning framework, which bypasses  assumptions on 
		data
		distributions and stationarity and replaces knowledge of second-order
		(cross-)covariances with knowledge of 	$\fullkernelspatiomat\timenot{\timeind}$ and
		$\fullkernelstatenoisemat\timenot{\timeind}$.
%	\begin{myitemize}
%		\myitem\cmt{kriging}by 
%		linear minimum mean-square error (LMMSE) estimation \cite[Ch. 
%		3]{cressie93}.
%		\myitem\cmt{krKF}To accommodate  time-evolving fields, \cite{mardia1998kriged} 
%		introduced 
%		the so-called
%		kriged Kalman filter (KriKF), which affords low-complexity online 
%		spatial prediction. A reduced-dimension version of the KriKF was introduced in 
%		\cite{wikle1999dimension} by expanding the spatio-temporal process as a linear 
%		combination 
%		of basis functions and applying the Kalman filter (KF) to the expansion coefficients.
%	\end{myitemize}
			\myitem\cmt{dynamic graphs}Moreover, different
			from~\cite{rajawat2014cartography,chua2006network,lorenzo2016lms,
				wang2015distributed}, the novel KeKriKF can accommodate 
				dynamic graph topologies	
				provided
				$\{\fullkernelspatiomat\timenot{\timeind} 
				,\fullkernelstatenoisemat\timenot{\timeind}\}_\timeind$ are available.
		\end{myitemize}
	\end{myitemize}
	
	\begin{algorithm}[t]                
		\caption{Kernel Kriged Kalman filter (\textbf{KeKriKF})}
		\label{algo:krkalmanfilter}    
		\begin{minipage}{40cm}
						\indent\textbf{Input:} 
			$\fullkernelstatenoisemat\timenot{\timeind}, 
			\fullkernelspatiomat\timenot{\timeind}\in\pdset^{\vertexnum}; $
			$\adjtransgraphmat\timetimenot{\timeind}{\timeind-1}\in 
			\rfield^{\vertexnum\times\vertexnum}$; 
			$\observationvec\timenot{\timeind}\in 
			\rfield^{\samplenum\timenot{\timeind}}$; \\
			\vspace{0.3cm}
			\indent\hspace{1cm} $\samplemat\timenot{\timeind}\in\{0,1\}^ 
			{\samplenum\timenot{\timeind}\times\vertexnum}$; 
			$\estsignalspatiotempcomp
			 \timegiventimenot{\timeind-1}{\timeind-1}\in\rfield^{\vertexnum}$; 
			 $
			\errormat\timegiventimenot{\timeind-1}{\timeind-1}\in 
			\pdset^{\vertexnum}$.
			\\
		%	\begin{algorithm}[1]
%				\begin{steps}
%						% \hspace{4.2 em}
%			\item
%						\end{steps}
		S1. 	$
			\genkernelmat\timenot{\timeind}=\frac{1}{\regpartwo} 
			\samplemat\timenot{\timeind}\fullkernelspatiomat\timenot{\timeind} 
			\samplemat\transpose\timenot{\timeind}+ 
			\samplenum\timenot{\timeind}\identitymat_
			{\samplenum\timenot{\timeind}}$\label{step:init}
	\\
			S2.  $\estsignalspatiotempcomp 
			\timegiventimenot{\timeind}{\timeind-1}=\adjtransgraphmat\timetimenot{\timeind}{\timeind-1}
			\estsignalspatiotempcomp\timegiventimenot{\timeind-1}{\timeind-1}$\hspace{7em} 
			(\emph{prediction})
			\label{step:pred}\\
			S3.	$
			\errormat\timegiventimenot{t}{t-1}=\adjtransgraphmat\timetimenot{\timeind}{\timeind-1} 
			\errormat
			\timegiventimenot{t-1}{t-1}\adjtransgraphmat\transpose\timetimenot{\timeind}{\timeind-1}+
			\frac{1}{\regparone}\fullkernelstatenoisemat\timenot{t}$\label{step:errorpred}
		   \\
			S4. $
			\kalmangainmat\timenot{t}
			=\errormat\timegiventimenot{t}{t-1} 
			\samplemat\transpose\timenot{t}(
			\genkernelmat\timenot{\timeind}
			+\samplemat\timenot{t}\errormat\timegiventimenot{t}{t-1}
			\samplemat\transpose\timenot{t})^{-1}$ (\emph{gain})\label{step:kalmgain}\\
			S5.	$  
			\errormat\timegiventimenot{t}{t}=(\identitymat-\kalmangainmat 
			\timenot{t}\samplemat\timenot{t})\errormat\timegiventimenot{t}{t-1}$\label{step:errorcor}\\
			S6.	 $  
			\estsignalspatiotempcomp\timenot{t|t}= 
			\estsignalspatiotempcomp\timenot{t|t-1}+		 
			\kalmangainmat\timenot{t}(\observationvec\timenot{t}- 
			\samplemat\timenot{t}\estsignalspatiotempcomp\timenot{t|t-1})
			$ \hspace{3em} (\emph{correction})\label{step:cor}
			\\
				S7. $
			\estsignalspatiocomp\timegiventimenot{\timeind}{\timeind}= 
			\fullkernelspatiomat\timenot{\timeind} 
			\samplemat\transpose\timenot{\timeind} 
			{\genkernelmat\timenot{\timeind}}\inv
			(\observationvec\timenot{\timeind} 
			-\samplemat\timenot{\timeind}\estsignalspatiotempcomp\timenot{\timeind|\timeind})
			$ \hspace{2em}  (\emph{kriging})\label{step:krig}
			\vspace{0.3cm}\\	
%			\end{algorithm}		
			\indent\textbf{Output:} 
			$\estsignalspatiotempcomp\timegiventimenot{\timeind}{\timeind}$;
			$\estsignalspatiocomp\timegiventimenot{\timeind}{\timeind}$; 
			$\errormat\timegiventimenot{\timeind}{\timeind}$.
			%\UNTIL{ $\|                          \auxveco\iternot{k+1} 
			%-                          
			%\trsamplealphavec\iternot{k+1}\| 
			%\leq                          \epsilon $ }
			%		\end{algorithmic}
		\end{minipage}
	\end{algorithm}
	
	\myitem\cmt{remark complexity}
	\begin{myremarkhere}\label{re:complex}
       The complexity of KeKriKF is $\mathcal{O}(\vertexnum^3)$ per slot. When  the underlying graph is 
       large ($\vertexnum\gg$), this complexity can be managed after splitting  
       the graph 
		into $N_g$ subgraphs each with at most $\lceil N/N_g\rceil$ nodes, and employing 
		consensus-based 
		decentralized KF schemes along the lines of ~\cite{schizas2008consensus}.
	\end{myremarkhere}
\end{myitemize}
\section{Online multi-kernel learning}
This section  broadens the scope  of the KeKriKF algorithm by employing a  multi-kernel 
learning scheme, to 
bypass the need for selecting an appropriate kernel. 

\label{sec:omk}
\begin{myitemize}
	\myitem\cmt{backround in MKL}The performance of KRR estimators is well 
	known 
	to
	heavily depend on 
	the choice of the kernel matrix~\cite{romero2016multikernel}.
	Unfortunately, it is difficult to know which kernel matrix is  most 
	appropriate for a given problem. 
	%In lieu of prior information about the 
	%suitable kernel motivates data-adaptive techniques that select the 
	%most appropriate kernel within a given pool.
	  To address this issue, 
	an MKL approach is presented that selects a suitable kernel matrix within 
	the linear 
	span of a
	prespecified dictionary  using the available data. 
	%Such an approach 
	%obviates the need for 
	%prior knowledge about the graph function 
    %structure.

	In the 
	following, consider for simplicity that 
	$\fullkernelspatiomat\timenot{\timeind}=\fullkernelspatiomat$,
	$\fullkernelstatenoisemat\timenot{\timeind}=\fullkernelstatenoisemat$, and 
	$\samplemat\timenot{\timeind}=\samplemat ,~\forall\timeind$. 
\cmt{specific}The kernels in the 
dictionaries 
$\fullkernelspatiomatdict\define\{\fullkernelspatiomat\kernelindnot{\rkhsind} 
\in\pdset^{\vertexnum}\}_{\rkhsind=1}^\rkhsspationum$, and 
$\fullkernelstatenoisematdict\define\{\fullkernelstatenoisemat\kernelindnot{\rkhsind}
\in\pdset^{\vertexnum}\}_{\rkhsind=1}^\rkhsstatenoisenum$  will be combined to 
generate
$\fullkernelspatiomat=
\fullkernelspatiomat\kernelcoefnot{\kernelcoefspatiovec} 
:= 
\sum_{\rkhsind=1}^{\rkhsspationum}\kernelcoefspatio\kernelindnot{\rkhsind} 
\fullkernelspatiomat\kernelindnot{\rkhsind}$ and 
$\fullkernelstatenoisemat=\fullkernelstatenoisemat 
\kernelcoefnot{\kernelcoefstatenoisevec}:= 
\sum_{\rkhsind=1}^{\rkhsstatenoisenum} 
\kernelcoefstatenoise\kernelindnot{\rkhsind} 
\fullkernelstatenoisemat\kernelindnot{\rkhsind}$, where 
$\kernelcoefspatiovec:=[\kernelcoefspatio\kernelindnot{1},\ldots, 
\kernelcoefspatio\kernelindnot{\rkhsspationum}]\transpose$, $ 
\kernelcoefstatenoisevec:=[\kernelcoefstatenoise\kernelindnot{1},\ldots, 
\kernelcoefstatenoise\kernelindnot{\rkhsstatenoisenum}]\transpose \succeq \bm  0 $ 
are coefficients to be determined.
% and
%$\fullkernelspatiomatdict\define\{\fullkernelspatiomat\kernelindnot{\rkhsind} 
%\in\pdset^{\vertexnum}\}_{\rkhsind=1}^\rkhsspationum$, 
%$\fullkernelstatenoisematdict\define\{\fullkernelstatenoisemat\kernelindnot{\rkhsind}
%\in\pdset^{\vertexnum}\}_{\rkhsind=1}^\rkhsstatenoisenum$ 
%are known dictionaries of 
%kernels  that allow for increased flexibility.
%to the 
%reconstruction algorithm. 
%ictionaries
% Moreover, this paper 
%postulates 
%%that the kernel matrices 
%form a conic combination to provide the pertinent kernel 
%matrices, i.e.

\myitem\cmt{opt 
	problem reform}Next,
% in 
%accordance with Sec.~\ref{sec:backmkl},
 consider expanding the optimization 
in~\eqref{eq:rkhsobj} to 
 obtain $\kernelcoefspatiovec, \kernelcoefstatenoisevec $ along with
$\{
\truesignalspatiotempcomp\timenot{\tau},
\truesignalspatiocomp\timenot{\tau}
%\truesignalspatiocompfun\timenot{\tau}\in\rkhs_\spatioind\timenot{\tau},\atop
%\truesignalspatiotempcompfun\timenot{\tau}\in\rkhs_\spatiotempind\timenot{\tau}
\}_{\tau=1}^\timeind$, as follows
				\begin{align}
				%	 \{\estsignalspatiocomp\timenot{\tau},
				%	 
				%\estsignalspatiotempcomp\timenot{\tau|\timeind}\}_{\tau=1}^	
				\label{eq:multrkhsobj}		
				\underset{\{
					\truesignalspatiotempcomp\timenot{\tau},
					\truesignalspatiocomp\timenot{\tau}
					%\truesignalspatiocompfun\timenot{\tau}\in\rkhs_\spatioind\timenot{\tau},\atop
					%\truesignalspatiotempcompfun\timenot{\tau}\in\rkhs_\spatiotempind\timenot{\tau}
					\}_{\tau=1}^\timeind, \atop \kernelcoefstatenoisevec\succeq 
					\bm  0 , 
					\kernelcoefspatiovec\succeq  \bm 0 }{\minimize}&
				~\tfrac{1}{\timeind}\sum_{\tau=1}^{\timeind}\tfrac{1}{\samplenum}\|\observationvec\timenot{\tau}
				-
				\samplemat\truesignalspatiotempcomp\timenot{\tau}
				-\samplemat\truesignalspatiocomp\timenot{\tau}	  
				\|^2 \nonumber\\ 
				&\hspace{-2cm}+\tfrac{\regparone}{\timeind}\sum_{\tau=1}^{\timeind}\|\truesignalspatiotempcomp\timenot{\tau}-
				~\adjtransgraphmat\timetimenot{\tau} 
				{\tau-1}\truesignalspatiotempcomp\timenot 
				{\tau-1}\|^2_{\fullkernelstatenoisemat\kernelcoefnot{\kernelcoefstatenoisevec}}
				\\
				&\hspace{-2cm}+
				\tfrac{\regpartwo}{\timeind}\sum_{\tau=1}^{\timeind} 
				\|\truesignalspatiocomp\timenot{\tau}\|^2_ 
				{\fullkernelspatiomat\kernelcoefnot{\kernelcoefspatiovec}}
				+\kernelcoefspatioreg\|\kernelcoefspatiovec\|_2^2
				+\kernelcoefstatenoisereg\|\kernelcoefstatenoisevec\|_2^2\nonumber
				\end{align}
where $\kernelcoefspatioreg, \kernelcoefstatenoisereg\ge0$ are regularization 
parameters. The solution to \eqref{eq:multrkhsobj} for each $\timeind$ will be 
denoted as $\{\estsignalspatiotempcomp\timegiventimenot{\tau}{\timeind}, 
\estsignalspatiocomp \timegiventimenot{\tau}
{\timeind}\}_{\tau=1}^{\tau=\timeind}\cup\{\kernelcoefstatenoisevecest\timenot{\timeind}, 
\kernelcoefspatiovecest\timenot{\timeind}\}$.
%The Laplacian kernel regularizers promote desirable properties of  $ 
%\{\truesignalspatiotempcomp\timenot{\tau},
%\truesignalspatiocomp\timenot{\tau}\}_{\tau=1}^\timeind$ by employing  
%$
%\fullkernelstatenoisemat\kernelcoefnot{\kernelcoefstatenoisevecest}, 
%\fullkernelspatiomat\kernelcoefnot{\kernelcoefspatiovecest}$.  
Here, the data-dependent $\{\kernelcoefstatenoisevecest\timenot{\timeind}, 
\kernelcoefspatiovecest\timenot{\timeind}\}$ select the kernel matrices that 
``best'' 
capture the data dynamics. 
%As a result, the estimates of $\{
%\truesignalspatiotempcomp\timenot{\tau},
%\truesignalspatiocomp\timenot{\tau}
%%\truesignalspatiocompfun\timenot{\tau}\in\rkhs_\spatioind\timenot{\tau},\atop
%%\truesignalspatiotempcompfun\timenot{\tau}\in\rkhs_\spatiotempind\timenot{\tau}
%\}_{\tau=1}^\timeind$ from~\eqref{eq:multrkhsobj} adapt to changes in the 
%spatio-temporal dynamics 
%of 
%$\signalvec\timenot{\timeind}$.
%%, that effect a ball constraint on $\kernelcoefspatiovec$ and 
%$\kernelcoefstatenoisevec$.
%, weighted
%by $\timeind$ to account for the first three terms that are growing over time. 
%	One can generalize the Laplacian kernel regularizers by 
%	introducing 
%	different 
%	parameters $\kernelcoefspatioreg\kernelindnot{\rkhsind} 
%	(\kernelcoefspatio\kernelindnot{\rkhsind})^2$ and 
%$\kernelcoefstatenoisereg\kernelindnot{\rkhsind} 
%	(\kernelcoefstatenoise\kernelindnot{\rkhsind})^2$ for each 
%	kernel in the dictionary.
%	In this way prior knowledge about the signal	 structure [cf. 
%	Table~\ref{tab:spectralweightfuns}] can be promoted. 
%	\cmt{tv estimates}The solution 
%	of~\eqref{eq:multrkhsobj} 
%	provides time-varying    estimates $ 
%	\kernelcoefspatiovecest\timenot{\timeind}$ and $ 
%	\kernelcoefstatenoisevecest\timenot{\timeind}$,
%	that allow tracking changes in the pertinent $\fullkernelspatiomat\timenot{\timeind}$ and 
%	$\fullkernelstatenoisemat\timenot{\timeind}$ and adapt to changes in the spatio-temporal dynamics 
%	of the graph 
%	functions.
	
\begin{myitemize}
\myitem\cmt{convexity analysis}Due to the presence of  the weighted norms, 
namely  
$\{\|\truesignalspatiotempcomp\timenot{\tau}-
~\adjtransgraphmat\timetimenot{\tau} {\tau-1}\truesignalspatiotempcomp\timenot 
{\tau-1}\|^2_{\fullkernelstatenoisemat\kernelcoefnot{\kernelcoefstatenoisevec}} 
\}_{\tau=1}^\timeind$
and $\{
\|\truesignalspatiocomp\timenot{\tau}\|^2_ 
{\fullkernelspatiomat\kernelcoefnot{\kernelcoefspatiovec}} 
\}_{\tau=1}^\timeind$, the problem in~\eqref{eq:multrkhsobj} is  non-convex.
  Fortunately,~\eqref{eq:multrkhsobj} is separately convex in $\{
\truesignalspatiotempcomp\timenot{\tau},
\truesignalspatiocomp\timenot{\tau}
%\truesignalspatiocompfun\timenot{\tau}\in\rkhs_\spatioind\timenot{\tau},\atop
%\truesignalspatiotempcompfun\timenot{\tau}\in\rkhs_\spatiotempind\timenot{\tau}
\}_{\tau=1}^\timeind, \kernelcoefspatiovec, 
\kernelcoefstatenoisevec$, which motivates the use of alternating minimization
(AM) strategies. AM algorithms minimize the  objective with respect to every 
block of
variables, while keeping the other variables fixed~\cite{csisz1984information}.
%and alternate over all the 
%variables until \acom{convergence to 
%a 
%stationary point}.% solution 
%of~\eqref{eq:multrkhsobj}. 
\myitem\cmt{approximate solution}Conveniently, if $\kernelcoefspatiovec, 
\kernelcoefstatenoisevec$ are fixed, then \eqref{eq:multrkhsobj} reduces to 
\eqref{eq:rkhsobj}, 
which can be solved by
Algorithm~\ref{algo:krkalmanfilter} for  
$\estsignalspatiocomp\timegiventimenot{\timeind}{\timeind}, 
\estsignalspatiotempcomp\timegiventimenot{\timeind}{\timeind}$ per slot
$\timeind$; see 
\thref{thm:kkrkf}. Conversely,  $\kernelcoefstatenoisevecest\timenot{\timeind}, 
\kernelcoefspatiovecest\timenot{\timeind}$ can be obtained for fixed 
$\{\truesignalspatiocomp\timenot{\tau},\truesignalspatiotempcomp
\timenot{\tau}\}_{\tau=1}^\timeind$ as specified next.

\begin{mytheoremhere}
	\thlabel{thm:mkrkf}Consider minimizing~\eqref{eq:multrkhsobj} with respect 
	to 
$\kernelcoefstatenoisevec$ and $ \kernelcoefspatiovec$ for fixed 
	$\truesignalspatiotempcomp 
	\timenot{\tau}=\estsignalspatiotempcomp\timegiventimenot 
	{\tau}{\tau}$ and
	$\truesignalspatiocomp\timenot{\tau}=\estsignalspatiocomp\timegiventimenot 
	{\tau}{\tau}$, $\tau=1,\ldots,\timeind$, where 
	$\{\estsignalspatiotempcomp\timegiventimenot{\tau}{\tau}, 
	\estsignalspatiocomp\timegiventimenot{\tau}{\tau}\}_ 
	{\tau=1}^{\timeind}$ are given and not necessarily the global 
	minimizers 
	of~\eqref{eq:multrkhsobj} with respect to $\{
	\truesignalspatiotempcomp\timenot{\tau},
	\truesignalspatiocomp\timenot{\tau}
	%\truesignalspatiocompfun\timenot{\tau}\in\rkhs_\spatioind\timenot{\tau},\atop
	%\truesignalspatiotempcompfun\timenot{\tau}\in\rkhs_\spatiotempind\timenot{\tau}
	\}_{\tau=1}^\timeind$.
	Let  $\residualstateervec 
	\timenot{\tau|\tau}\define\estsignalspatiotempcomp\timenot{\tau|\tau}
	-\adjtransgraphmat\timetimenot{\tau}{\tau-1}\estsignalspatiotempcomp\timenot
	 {\tau-1|\tau-1},~\tau=2,\ldots,\timeind$,
	 as well as
	$\spatiocorsmat\timenot{\timeind}
	=\tfrac{1}{\timeind}\sum_{\tau=1}^{\timeind}\estsignalspatiocomp\timegiventimenot{\tau}{\tau}
	{\estsignalspatiocomp\timegiventimenot{\tau}{\tau}}\transpose$ and 
	$\statenoisecorsmat\timenot{\timeind}%\timetotimenot{1}{\timeind}
	=\tfrac{1}{\timeind}\sum_{\tau=1}^{\timeind}\residualstateervec\timegiventimenot{\tau}{\tau}
	{\residualstateervec\timegiventimenot{\tau}{\tau}}\transpose$.
	%The following optimization problems solve~\eqref{eq:multrkhsobj} for
	 Then, the minimizers of \eqref{eq:multrkhsobj} with respect to 
	$\kernelcoefspatiovec$ and 
	$ 
	\kernelcoefstatenoisevec$ are 
		\begin{subequations}
				\label{eq:multkerwithcor}
		\begin{align}
		\label{eq:spatiomultkerwithcor}
		%	\kernelcoefspatiovecest\timenot{\timeind}&=\underset{\kernelcoefspatiovec\ge
		%	 \bm 0}{\argmin} \Tr{(\samplespatiocormat\timenot{\timeind}\genkernelmat\inv
		%	 \kernelcoefnot{\kernelcoefspatiovec})}
		%	+\kernelcoefspatioreg\|\kernelcoefspatiovec\|_2^2\\
		\hspace{-0.2cm}\kernelcoefspatiovecest\timenot{\timeind}&=\underset{\kernelcoefspatiovec\succeq
			\bm 0}{\argmin} \Tr\{\spatiocorsmat\timenot{\timeind}{\fullkernelspatiomat}\inv
			\kernelcoefnot{\kernelcoefspatiovec}\}
		+\tfrac{\kernelcoefspatioreg}{\regpartwo}\|\kernelcoefspatiovec\|_2^2\\
		\hspace{-0.2cm}\kernelcoefstatenoisevecest\timenot{\timeind}&=\underset{\kernelcoefstatenoisevec\succeq
			\bm 
			0}{\argmin}
		\Tr\{\statenoisecorsmat\timenot{\timeind}{\fullkernelstatenoisemat}\inv 
			\kernelcoefnot{\kernelcoefstatenoisevec}\}
		+\tfrac{\kernelcoefstatenoisereg}{\regparone}\| 
		\kernelcoefstatenoisevec\|_2^2.
		\label{eq:stateermultkerwithcor}
		\end{align}
	\end{subequations}
%	\begin{subequations}
%	\begin{align}
%	\label{eq:spatiomultkerwithcor}
%	%	\kernelcoefspatiovecest\timenot{\timeind}&=\underset{\kernelcoefspatiovec\ge
%	%	 \bm 0}{\argmin} \Tr{(\samplespatiocormat\timenot{\timeind}\genkernelmat\inv
%	%	 \kernelcoefnot{\kernelcoefspatiovec})}
%	%	+\kernelcoefspatioreg\|\kernelcoefspatiovec\|_2^2\\
%	\kernelcoefspatiovecest\timenot{\timeind}&=\underset{\kernelcoefspatiovec\ge
%		\bm 0}{\argmin} \Tr{(\spatiocormat\timenot{\timeind}\fullkernelspatiomat\inv
%		\kernelcoefnot{\kernelcoefspatiovec})}
%	+\kernelcoefspatioreg\|\kernelcoefspatiovec\|_2^2\\
%	\kernelcoefstatenoisevecest\timenot{\timeind}&=\underset{\kernelcoefstatenoisevec\ge
%		\bm 
%		0}{\argmin}
%	\Tr{(\statenoisecormat\timenot{\timeind}\fullkernelstatenoisemat\inv 
%		\kernelcoefnot{\kernelcoefstatenoisevec})}
%	+\kernelcoefstatenoisereg\|\kernelcoefstatenoisevec\|_2^2
%	\label{eq:stateermultkerwithcor}
%	\end{align}
%\end{subequations}
\end{mytheoremhere}
\noindent\emph{Proof:} To prove~\eqref{eq:spatiomultkerwithcor}, keep 
in~\eqref{eq:multrkhsobj}  only those terms that depend on 
$\kernelcoefspatiovec$,
 and replace $\{\truesignalspatiocomp\timenot{\tau}
\}_{\tau=1}^\timeind$ with
$\{\estsignalspatiocomp\timegiventimenot{\tau}{\tau}\}_{\tau=1}^\timeind$. 
Then, the 
objective in~\eqref{eq:multrkhsobj}  
reduces to
$({1}/{\timeind})\sum_{\tau=1}^{\timeind}{\estsignalspatiocomp
\timegiventimenot{\tau}{\tau}}\transpose {\fullkernelspatiomat}\inv
			\kernelcoefnot{\kernelcoefspatiovec}\estsignalspatiocomp\timegiventimenot{\tau}{\tau}
		+({\kernelcoefspatioreg}/{\regpartwo})\|\kernelcoefspatiovec\|_2^2$. Next, using the linearity 
		and cyclic invariance of 
		the trace it follows that
		$\Tr\big\{({{1}/{\timeind})\sum_{\tau=1}^{\timeind} 
		{\estsignalspatiocomp\timegiventimenot{\tau}{\tau}}\transpose 
		{\fullkernelspatiomat}\inv
		\kernelcoefnot{\kernelcoefspatiovec} 
		\estsignalspatiocomp\timegiventimenot{\tau}{\tau}}\big\}=$ $
	\Tr\big\{({1}/{\timeind})\sum_{\tau=1}^{\timeind} 
		\estsignalspatiocomp\timegiventimenot{\tau}{\tau} 
		{\estsignalspatiocomp\timegiventimenot{\tau}{\tau}}\transpose{\fullkernelspatiomat}\inv
	\kernelcoefnot{\kernelcoefspatiovec}\big\}= 
	\Tr\big\{\spatiocorsmat\timenot{\timeind}{\fullkernelspatiomat}\inv
		\kernelcoefnot{\kernelcoefspatiovec}\big
		\}$, which proves 
		\eqref{eq:spatiomultkerwithcor}. The proof of 
		\eqref{eq:stateermultkerwithcor} follows along the same lines. \qed

\end{myitemize}

\myitem\cmt{General optimization problem}Thus, \thref{thm:mkrkf} simplifies the objective that has to 
be minimized to find $\kernelcoefstatenoisevecest\timenot{\timeind}$ and
$\kernelcoefspatiovecest\timenot{\timeind}$. With 
$\fullkernelmat\kernelcoefnot{\kernelcoefvec}= 
\sum_{\rkhsind=1}^{\rkhsnum}\kernelcoef\kernelindnot{\rkhsind} 
\fullkernelmat\kernelindnot{\rkhsind}\nonumber$, problems 
\eqref{eq:spatiomultkerwithcor} and \eqref{eq:stateermultkerwithcor} are of the 
form
\begin{align}
\kernelcoefvecest=\underset{\kernelcoefvec\ge
	\bm 
	0}{\argmin}&~%\optobjfun\optobjno{\kernelcoefvec}\define
\Tr\{\corrmat\fullkernelmat\inv 
	\kernelcoefnot{\kernelcoefvec}\}
+\kernelcoefreg\|\kernelcoefvec\|_2^2
\label{eq:genmultkerwithcor}
\end{align}
 for some $\corrmat\in\rfield^{\vertexnum\times\vertexnum}$, 
 $\kernelcoefreg\ge0$, and 
 $\kernelmatdict=\{\fullkernelmat\kernelindnot{\rkhsind} 
\}_{\rkhsind=1}^\rkhsnum$.
	\myitem\cmt{kernel matching}Due to their resemblance to \emph{covariance 
matching}~\cite{romero2013wideband},  problem \eqref{eq:genmultkerwithcor}, 
and hence \eqref{eq:spatiomultkerwithcor} and \eqref{eq:stateermultkerwithcor} 
will be referred to as \emph{kernel matching}.

\thref{thm:mkrkf} suggests an online AM procedure to approximate the 
solution to~\eqref{eq:multrkhsobj}, where Algorithm~\ref{algo:krkalmanfilter} 
and a solver for~\eqref{eq:genmultkerwithcor} termed online kernel matching (OKM)
% that will be described later 
 are executed 
alternatingly. This is summarized as 
Algorithm~\ref{algo:mkkrkf}, and it is termed multi-kernel KriKF (MKriKF).
%is a solver for~\eqref{eq:genmultkerwithcor}.that will be described later 
%and hence for 
%\eqref{eq:multrkhsobj}, . Note that 
Algorithm~\ref{algo:mkkrkf} does not generally find a global optimum of 
\eqref{eq:multrkhsobj}; yet, finding such an optimum may not be critical in 
practice, since it cannot be computed in polynomial time.

The rest of this section develops the OKM algorithm for 
solving~\eqref{eq:genmultkerwithcor} when $\kernelmatdict$ comprises Laplacian 
kernels. The first step is to exploit the fact that all Laplacian kernel 
matrices associated with a given 
graph have common eigenvectors.
\end{myitemize}
\begin{myitemize}

	\myitem\cmt{Solution PGD}
%Since \eqref{eq:genmultkerwithcor} is strongly convex an efficient algorithm will be derived based on 
%projected gradient descent (PGD).
%The ensuing propositions will come handy in developing efficient 
%solvers of~\eqref{eq:genmultkerwithcor}. 
\begin{myitemize}
	\myitem\cmt{reform}\emph{\begin{proposition}\label{prop:ref}
	    Consider the
%		Let $\{\}_{\rkhsind=1}^\rkhsnum$
%		have 
eigenvalue 		decompositions 
		$\{\fullkernelmat\kernelindnot{\rkhsind}=
		\fullkernelevecmat\diag{\laplacianevalvec\kernelindnot{\rkhsind}} 
		\fullkernelevecmat\transpose\}_{\rkhsind=1}^\rkhsnum$ and let  
		$\corrmatprojected\define 
		\fullkernelevecmat\transpose\corrmat\fullkernelevecmat$.
		%, where 
		%$\laplacianmat=\fullkernelevecmat \diag{\laplacianevalvec} 
		%\fullkernelevecmat\transpose$.
		 Upon defining $\fullkernelevalmat\kernelcoefnot{\kernelcoefvec}	
		 \define\diag{\sum_{\rkhsind=1} 
		 	^{\rkhsnum}\kernelcoef\kernelindnot{\rkhsind} 
		 	\laplacianevalvec\kernelindnot 
		 	{\rkhsind}}$ and $\optobjfun\optobjno{\kernelcoefvec}\define
		 \Tr{(\corrmatprojected\fullkernelevalmat\inv 
		 	\kernelcoefnot{\kernelcoefvec})}
		 +\kernelcoefreg\|\kernelcoefvec\|_2^2$,~\eqref{eq:genmultkerwithcor}  
		 can be equivalently 
		 	written as
		\begin{align}
			\kernelcoefvecest=\underset{\kernelcoefvec\succeq
				\bm 
				0}{\argmin}&~~\optobjfun\optobjno{\kernelcoefvec}
			\label{eq:genmultkereigwithcor}
		\end{align}
		\end{proposition}}
	\noindent\emph{Proof:} 
			Since 
			$\fullkernelmat\kernelcoefnot{\kernelcoefvec}= 
			\sum_{\rkhsind}^{\rkhsnum}\kernelcoef\kernelindnot{\rkhsind} \fullkernelevecmat 
			\diag{\laplacianevalvec\kernelindnot{\rkhsind}}\fullkernelevecmat\transpose$ $= 
			\fullkernelevecmat\fullkernelevalmat\kernelcoefnot{\kernelcoefvec} 
			\fullkernelevecmat\transpose$, 
			\eqref{eq:genmultkereigwithcor} follows by noting that
			$\Tr\{\corrmat\fullkernelmat\inv 
			\kernelcoefnot{\kernelcoefvec}\}=\Tr\{\corrmat\fullkernelevecmat 
			\fullkernelevalmat\inv \kernelcoefnot{\kernelcoefvec}
			\fullkernelevecmat\transpose\}=\Tr\{
			\fullkernelevecmat\transpose\corrmat\fullkernelevecmat
			\fullkernelevalmat\inv \kernelcoefnot{\kernelcoefvec}\}$ $=\Tr\{
	\corrmatprojected
	\fullkernelevalmat\inv \kernelcoefnot{\kernelcoefvec}\}$.
\qed

	Proposition~\ref{prop:ref} establishes that \eqref{eq:genmultkerwithcor} 
	can be expressed as \eqref{eq:genmultkereigwithcor} when the kernels in 
	$\kernelmatdict$ share eigenvectors, as is the case of Laplacian 
	kernels; cf. Sec.~\ref{sec:background}.
	\myitem\cmt{grad calculation}
	\emph{\begin{proposition}\label{prop:grad}
   When $\kernelcoefvec\succeq\bm0$, function 
   $\optobjfun\optobjno{\kernelcoefvec}$ is strongly convex and differentiable 
   with 
   gradient
   \begin{align}
   	\label{eq:grad}
   	\nabla\optobjfun\optobjno{\kernelcoefvec}=\gradtrvec
   	\kernelcoefnot{\kernelcoefvec}	
   	+2\kernelcoefreg	
   	\kernelcoefvec
   \end{align}
   where $\gradtrvec\kernelcoefnot{\kernelcoefvec}	
   \define-[\Tr\big\{\text{diag}\{\fullkernelevaltrvec
   	\kernelindnot{1}\}\corrmatprojected\big\},\ldots,$ 
   	$\Tr\big\{\text{diag}\{\fullkernelevaltrvec
   	\kernelindnot{\rkhsnum}\}\corrmatprojected\big\}]$, with 
   	$\fullkernelevaltrvec
\kernelindnot{\rkhsind}\define[\fullkernelevaltr\vertexnot{1} 
\kernelindnot{\rkhsind}, 
\ldots,\fullkernelevaltr\vertexnot{\vertexnum}\kernelindnot{\rkhsind}]\transpose$ and 
$\fullkernelevaltr\vertexnot{\vertexind}\kernelindnot{\rkhsind}\define
{\laplacianeval\vertexnot{\vertexind}\kernelindnot{\rkhsind}}/ 
{({\sum_{\mu=1}^{\rkhsnum}}\kernelcoef\kernelindnot{\mu}\laplacianeval\vertexnot{\vertexind} 
\kernelindnot{\mu})^2}$.
	\end{proposition}}
\noindent\emph{Proof:} Because $\corrmatprojected$ is a 
		positive semidefinite matrix and 
		$\laplacianevalvec\kernelindnot{\rkhsind}\succeq\bm{0}~\forall\rkhsind$, it can be 
		easily seen that  $\Tr\big\{
		\corrmatprojected
		\fullkernelevalmat\inv \kernelcoefnot{\kernelcoefvec}\big\}$ is convex 
		over 
		$\kernelcoefvec\succeq\bm{0}$. And since 
		$\kernelcoefreg\|\kernelcoefvec\|_2^2$ is strongly convex, it follows  
		by its definition that $\optobjfun\optobjno{\kernelcoefvec}$ is strongly 
		convex.
		To obtain the gradient observe that \begin{align}
		\frac{\partial \optobjfun}{\partial \kernelcoef\kernelindnot{\rkhsind} }= 
		-\Tr\big\{\fullkernelevalmat\inv 
		\kernelcoefnot{\kernelcoefvec}\diag{\laplacianevalvec\kernelindnot 
		{\rkhsind}} 
		\fullkernelevalmat\inv
		\kernelcoefnot{\kernelcoefvec}\corrmatprojected\big\}+2\kernelcoefreg	
		\kernelcoef\kernelindnot{\rkhsind}
		\end{align}
		and $\fullkernelevalmat\inv 
		\kernelcoefnot{\kernelcoefvec}\diag{\laplacianevalvec\kernelindnot{\rkhsind}}
		\fullkernelevalmat\inv 
		\kernelcoefnot{\kernelcoefvec}=\text{diag}\{\fullkernelevaltrvec
			\kernelindnot{\rkhsind}\}
%		=
%		\diag{\laplacianevalvec\kernelindnot{\rkhsind}}
%		\diag{\sum_{\mu}^{\rkhsnum}\kernelcoef\kernelindnot{\mu} 
%			\laplacianevalvec\kernelindnot{\mu}}^{-2}
		$. 
\qed

\myitem\cmt{pgd}As \eqref{eq:genmultkereigwithcor} entails a strongly convex 
and 
differentiable objective, and projections on its feasible set are easy to 
obtain, we are motivate to solve \eqref{eq:genmultkereigwithcor} through 
projected gradient descent (PGD)~\cite{bertsekas1999nonlinear}. Besides its 
simplicity, PGD converges linearly to the 
	global minimum of~\eqref{eq:genmultkereigwithcor}. The general PGD 
	iteration is
%\begin{align}
%	\label{eq:gradcormatch}
%	\nabla_{\kernelcoefvec}\optobjfun\optobjno{\kernelcoefvec}=\gradtrvec+2\kernelcoefreg	
%	\kernelcoefvec
%\end{align}
% 	\myitem\cmt{PGD}The global optimum for \eqref{eq:genmultkerwithcor} is found by 
% 	PGD~\cite{bertsekas1999nonlinear}\acom{move earlier} as follows
 	\begin{align}
 	\label{eq:gradprojdec}
 	\kernelcoefvec\itergdnot{\itergd+1}=\posprojnot{\kernelcoefvec\itergdnot{\itergd}- 
 		\stepsize\itergdnot{\itergd}\nabla
 		\optobjfun\optobjno{\kernelcoefvec\itergdnot{\itergd}}},~\itergd=0,1,\ldots
 	\end{align}
 	where 
 	$\stepsize\itergdnot{\itergd}$ is the stepsize chosen e.g. by the Armijo 
 	rule~\cite{bertsekas1999nonlinear}, 
 	$\kernelcoefvec\itergdnot{0}$ is a feasible initial step,
 	and 
 	$\posprojnot{\cdot}$ denotes projection on the non-negative 
 	orthant $\{\kernelcoefvec: 
 	\kernelcoef\kernelindnot{\rkhsind}\ge0,~\rkhsind=1,\ldots,\rkhsnum\}$. 
 	The overall algorithm is termed OKM, and it is listed 
 	as 
 	Algorithm~\ref{algo:onlinemult}.
 
 		 \myitem\cmt{PGD for online case}Observe 
 		 %from 		 Algorithms~\ref{algo:mkkrkf} and \ref{algo:onlinemult}
 		  that $\kernelcoefvec\itergdnot{0}$ in Algorithm~\ref{algo:onlinemult} is 
 		 initialized with the output of Algorithm~\ref{algo:onlinemult}  in 
 		 the 
 		 previous iterate, namely $\kernelcoefvecest\timenot{\timeind-1}$. This 
 		 is a warm start 
 		 that considerably speeds up convergence of  
 		 Algorithm~\ref{algo:onlinemult}  since  
 		 $\optobjfun\optobjno{\kernelcoefvec}$
 		 is expected to change slowly across the iterations in 
 		 Algorithm~\ref{algo:mkkrkf}. An interesting 
 		 byproduct of 
 		 the OKM algorithm
 		 %adopted online AM approach in Algorithm~\ref{algo:mkkrkf}
 		  is its ability to adapt to 
 		 changes in the spatio-temporal dynamics of the graph functions by 
 		 adjusting the coefficients 
 		 $\{\kernelcoefspatiovecest\timenot{\timeind}, 
 		 \kernelcoefstatenoisevecest\timenot{\timeind}\}_\timeind$,
 		 and consequently the kernel matrices.
 		 
 		 \myitem\cmt{complexity}In view of Proposition~\ref{prop:grad}, 
 		 finding
 		 each entry of $\nabla\optobjfun\optobjno{\kernelcoefvec}$ in Algorithm~\ref{algo:onlinemult} 
 		 requires 
 		 $\mathcal{O}(\vertexnum)$ operations.
 		 %complexity since $\Tr(\diag{\fullkernelevaltrvec
 		 %	
 		 %\kernelindnot{\rkhsind}}\corrmatprojected)=\sum_{\vertexind=1}^{\vertexnum}
 		 %	
 		 %\fullkernelevaltr\vertexnot{\vertexind}\kernelindnot
 		 %{\rkhsind}\corrmatprojectedentry 
 		 %\vertexvertexnot{\vertexind}{\vertexind}$. 
 		 Computing the gradient through~\eqref{eq:grad} exploits 
 		 the common eigenvectors of 
 		 $\{\fullkernelmat\kernelindnot{\rkhsind}\}_{\rkhsind=1}^\rkhsnum$, and 
 		 avoids the 
 		 inversion of the 
 		 $\vertexnum\times\vertexnum$ matrix
 		 $\fullkernelmat\kernelcoefnot{\kernelcoefvec}$  
 		 that is 
 		 required when 
 		 calculating the gradient for the general 
 		 formulation~\eqref{eq:genmultkerwithcor}, where 
 		 $\{\fullkernelmat\kernelindnot{\rkhsind}\}_{\rkhsind=1}^\rkhsnum$ need not share eigenvectors.
 		 %see e.g.~\cite{zhang2016multikernelssp,cortes2009learning}.
 		 The 
 		 complexity of 
 		 evaluating 
 		 the 
 		 gradient is therefore reduced from a prohibitive 
 		 $\mathcal{O}(\vertexnum^{3}\rkhsnum)$ for general kernels to an 
 		 affordable 
 		 $\mathcal{O}(\vertexnum\rkhsnum)$ for Laplacian kernels, which amounts 
 		 to considerable computational savings especially for 
 		 large-scale 
 		 networks.  With $\maxiternumgd$ denoting the number of PGD iterations 
 		 for 
 		 convergence, the overall computational complexity of OKM is therefore
 		 $\mathcal{O}(\vertexnum\rkhsnum\maxiternumgd)$.
 		 %, where the first term is 
 		 %attributed to calculating the matrix product $\corrmatprojected\timenot{\timeind}=
 		 %\fullkernelevecmat\transpose\corrmat\timenot{\timeind}\fullkernelevecmat$
 		 % and the second 
 		 %is due to the PGD algorithm. 
 		 Typically, $\vertexnum^3\ge\vertexnum\rkhsnum\maxiternumgd$ and hence 
% 		 the 
% 		 effective complexity of Algorithm~\ref{algo:onlinemult} is 
% 		 $\mathcal{O}(\vertexnum^3)$, which coincides with that of KeKriKF. Thus,
 		 the complexity of 
 		 Algorithm~\ref{algo:mkkrkf} is $\mathcal{O}(\vertexnum^3)$, while 
 		 learning the appropriate linear combination of kernels through 
 		 MKL does not increase the complexity order that can be further reduced as suggested in 
 		 Remark~\ref{re:complex}. 
 		 %Actually, M<N(N-1)/2 because then the kernels in the basis are linearly independent
 		 %Notice that since the estimates  
 		 %$\kernelcoefvecest\timenot{\timeind}$ change over time, the combined kernel matrix 
 		 %$\fullkernelmat\timenot{\timeind}$ changes as well.
\end{myitemize}
\end{myitemize}

\begin{algorithm}[t]                
	\caption{Multi-kernel KriKF (\textbf{MKriKF})%\acom{update style like Algo 1}
		}
	\label{algo:mkkrkf}    
	\begin{minipage}{30cm}
		\indent\textbf{Input:} 
		$\fullkernelspatiomatdict$;
		$\fullkernelstatenoisematdict$; 
		$\laplacianmat=\laplacianevecmat\transpose\diag{\laplacianevalvec}\laplacianevecmat$.
		\vspace{0.3cm}
		\begin{algorithmic}[1]	
		\STATE Initialize: $\kernelcoefspatiovecest\timenot{0}= 
		\kernelcoefstatenoisevecest\timenot{0}=[1,0,\ldots,0]$, 
		$\estsignalspatiotempcomp\timenot{0|0}=\bm 0$, 
		\\
		 \indent\hspace{1.3cm} 
		$
		 \errormat\timegiventimenot{0}{0}= 
		 \frac{1}{\regparone}\fullkernelstatenoisemat\kernelindnot{1}$, 
		 	\\
		 	\indent\hspace{1.3cm} 
		 $\laplacianevalvecspatio\kernelindnot{\rkhsind}
		 :=\diag{\laplacianevecmat\fullkernelspatiomat\kernelindnot{\rkhsind} 
		 \laplacianevecmat\transpose}
		  \forall\rkhsind$,	\\
		  \indent\hspace{1.3cm}  $\laplacianevalvecspatiotemp\kernelindnot{\rkhsind}
		  :=\diag{\laplacianevecmat\fullkernelstatenoisemat\kernelindnot{\rkhsind} 
		  	\laplacianevecmat\transpose}
		  \forall\rkhsind$.\\	
		\STATE \textbf{for}~{$\timeind=1,2,\ldots$}{  }\textbf{do}\\		
		\STATE\indent\hspace{0.3cm}\textbf{Input:} 
		$\adjtransgraphmat\timetimenot{\timeind}{\timeind-1}\in 
		\rfield^{\vertexnum\times\vertexnum}$; 
		$\observationvec\timenot{\timeind}\in 
		\rfield^{\samplenum\timenot{\timeind}}$; 
		$\samplemat\timenot{\timeind}\in\{0,1\}^ 
		{\samplenum\timenot{\timeind}\times\vertexnum}$.
		\vspace{0.1cm}
		\\
		\STATE \indent\hspace{0.3cm}$\fullkernelspatiomat\timenot{\timeind}=
		\fullkernelspatiomat\kernelcoefnot
		{\kernelcoefspatiovecest\timenot{\timeind}}$
		\\\STATE \indent\hspace{0.3cm}$\fullkernelstatenoisemat 
		\timenot{\timeind}=\fullkernelstatenoisemat 
		\kernelcoefnot{\kernelcoefstatenoisevecest\timenot{\timeind}}$
		\\
		\STATE \indent\hspace{0.3cm}$\{\estsignalspatiocomp
		\timegiventimenot{\timeind}{\timeind}, 
		\estsignalspatiotempcomp\timegiventimenot{\timeind}{\timeind}\}=
		\textbf{\kkrkf}(\fullkernelstatenoisemat\timenot{\timeind-1}, 
		\fullkernelspatiomat\timenot{\timeind-1}, 
		\adjtransgraphmat\timetimenot{\timeind}{\timeind-1},
		$\\
		 \indent\hspace{2.5cm}$\observationvec\timenot{\timeind}, 
		\samplemat\timenot{\timeind},
		\estsignalspatiotempcomp 
		\timegiventimenot{\timeind-1}{\timeind-1}, 
		\errormat\timegiventimenot{\timeind-1}{\timeind-1})$\\
		\STATE \indent\hspace{0.3cm}Update  $\spatiocorsmat\timenot{\timeind}$ and 
		$\statenoisecorsmat\timenot{\timeind}$\\
		\STATE\indent\hspace{0.3cm}$\corrmatprojectedspatio\timenot{\timeind}=	
		\fullkernelevecmat\transpose\spatiocorsmat\timenot{\timeind}\fullkernelevecmat$\\
		\STATE\indent\hspace{0.3cm}$\corrmatprojectedspatiotemp\timenot{\timeind}=
		\fullkernelevecmat\transpose\statenoisecorsmat\timenot{\timeind}\fullkernelevecmat$\\
		\STATE \indent\hspace{0.3cm}%$\fullkernelspatiomat\timenot{\timeind}=$\indent
		$\kernelcoefspatiovecest\timenot{\timeind}=\textbf{\omk}(\{\laplacianevalvecspatio 
		\kernelindnot{\rkhsind}\}_{\rkhsind=1}^\rkhsspationum,
		\corrmatprojectedspatio\timenot{\timeind},
		\kernelcoefspatiovecest\timenot{\timeind-1})$
		\\
		\STATE \indent\hspace{0.3cm}%$\fullkernelstatenoisemat\timenot{\timeind}=$ 
		$\kernelcoefstatenoisevecest\timenot{\timeind}=\textbf{\omk}(\{\laplacianevalvecspatiotemp 
		\kernelindnot{\rkhsind}\}_{\rkhsind=1}^\rkhsstatenoisenum,
		\corrmatprojectedspatiotemp\timenot{\timeind}, 
		\kernelcoefstatenoisevecest\timenot{\timeind-1})$
		\\			
		\vspace{0.1cm}	
		\STATE \indent\hspace{0.3cm}\textbf{Output:} 
		$\estsignalspatiotempcomp\timegiventimenot{\timeind}{\timeind}$;
		$\estsignalspatiocomp\timegiventimenot{\timeind}{\timeind}$; 
		$\errormat\timegiventimenot{\timeind}{\timeind}$.\\
		\STATE \textbf{end for}
		%\UNTIL{ $\|                          \auxveco\iternot{k+1} 
		%-                          
		%\trsamplealphavec\iternot{k+1}\| 
		%\leq                          \epsilon $ }
		%		\end{algorithmic}
		\end{algorithmic}
	\end{minipage}
\end{algorithm}

		\begin{algorithm}[t]                
			\caption{Online kernel matching (\textbf{OKM})}
			\label{algo:onlinemult}    
			\begin{minipage}{30cm}
			\indent\textbf{Input:} 
				$\{\laplacianevalvec\kernelindnot{\rkhsind} 
				\}_{\rkhsind=1}^\rkhsnum$;
				$\corrmatprojected\timenot{\timeind}\in\pdset^{\vertexnum}$;
				$\kernelcoefvecest\timenot{\timeind-1}
				\in\rfield_+^ 
				{\rkhsnum}$.
				\vspace{0.3cm}
				%;				$\laplacianevecmat\in\rfield^{\vertexnum\times\vertexnum}$.
				\begin{algorithmic}[1]
					\STATE Initialize:		$\kernelcoefvec\itergdnot{0}=\kernelcoefvecest\timenot{\timeind-1}$, 
				%	$\corrmatprojected= 
					%\fullkernelevecmat\transpose\corrmat\fullkernelevecmat
					%$
					\\
						\STATE		\textbf{while} stopping\_criterion not met 
								\textbf{do}\\
			\STATE		\indent\hspace{0.3cm}$	
				\kernelcoefvec\itergdnot{\itergd+1}=\posprojnot{\kernelcoefvec\itergdnot{\itergd}- 
					\stepsize\itergdnot{\itergd}\nabla\optobjfun\optobjno{\kernelcoefvec\itergdnot{\itergd}}}$
				\\\STATE				\indent\hspace{0.3cm}$\itergd\leftarrow\itergd+1$\\
					\STATE			\indent\textbf{end while}
											\end{algorithmic}
											\vspace{0.3cm}
				\indent\textbf{Output:} $\kernelcoefvec\itergdnot{\itergd}$.
				%\UNTIL{ $\|                          \auxveco\iternot{k+1} -                          
				%\trsamplealphavec\iternot{k+1}\| 
				%\leq                          \epsilon $ }
				%		\end{algorithmic}
			\end{minipage}
		\end{algorithm}

%\myitem\cmt{general algo}Finally, Algorithm~\ref{algo:mkkrkf} describes 
% the proposed multi-kernel KriKF (MKriKF) algorithm that  operates
% in an 
% online and 
% data-adaptive fashion. MKriKF alternates between computing  
% $\estsignalspatiotempcomp\timegiventimenot{\timeind}{\timeind}$ 
%, $\estsignalspatiocomp\timegiventimenot{\timeind}{\timeind}$ from 
%Algorithm~\ref{algo:krkalmanfilter} and  $ 
%\kernelcoefspatiovecest\timenot{\timeind}$ , $ 
%\kernelcoefstatenoisevecest\timenot{\timeind}$ from 
%Algorithm~\ref{algo:onlinemult}.
%\change{Notice that the computational complexity of  MKriKF, that in addition 
%adapts to the observed 
%data,  is 
%of 
%the 
%same order as KeKriKF at
%$\mathcal{O}(\vertexnum^3)$.}

%  and 
% alternates between estimation of $\rkhsvec\timenot{\timeind}$ and $\kernelcoefspatiovec, 
% \kernelcoefspatiovec$.
% \acom{Remark 7: (Convergence analysis) The convergence 
% analysis
% 	for Algorithm 2 is challenging for arbitrary observation windows.
% 	The main difficulty stems from the dependency of (23) on
% 	all previous observations .} 
% The algorithm is initialized with
% $	\fullkernelspatiomat\timenot{0}=\fullkernelspatiomat\kernelindnot{1}$,
% 
%$\fullkernelstatenoisemat\timenot{0}=\fullkernelstatenoisemat\kernelindnot{1}$,
% $\estsignalspatiotempcomp\timenot{0|0}=\bm 0$, and $
% \errormat\timegiventimenot{0}{0}= 
% \frac{1}{\regparone}\fullkernelstatenoisemat\kernelindnot{1}$.

\begin{myremarkhere}
	The algorithms in this section adopted a fixed kernel 
	dictionary over time, namely
	$\kernelmatdict=\{\fullkernelmat\kernelindnot{\rkhsind} 
	\in\pdset^{\vertexnum}\}_{\rkhsind=1}^\rkhsnum$. If the topology changes over time, the Laplacian 
	kernel 
	matrices change as
	well, cf. ~\eqref{eq:laplaciankerneldef}.  To accommodate this 
	scenario,  
	one 
	can restart Algorithm~\ref{algo:mkkrkf} whenever the 
	topology changes, say at time $\timetopologychange$, and initialize 
	$\estsignalspatiotempcomp\timenot{0|0}\leftarrow\estsignalspatiotempcomp\timenot{ 
	\timetopologychange|\timetopologychange}$, $
	\errormat\timegiventimenot{0}{0}\leftarrow\errormat\timegiventimenot{\timetopologychange} 
	{\timetopologychange}
	 $, as well as replace the Laplacian kernels in $\kernelmatdict$ with the 
	 ones corresponding to the new topology.
\end{myremarkhere}

	\begin{myremarkhere}
		%For alternative way to compute R see remark 7.\acom{should it be a 
		%remark?}
		\begin{myitemize}
			\myitem\cmt{non-stationary}
			To accommodate a certain degree of nonstationarity one may consider 
			using 
			the following matrices 
			\begin{subequations}
				\label{eq:rescor}
				\begin{align}
				\spatiocormat\timenot{\timeind}%\timetotimenot{1}{\timeind}
				=&\sum_{\tau=1}^{\timeind}
				\forgetingfactorspatio^{\timeind-\tau}
				\estsignalspatiocomp\timegiventimenot{\tau}{\tau}
				{\estsignalspatiocomp\timegiventimenot{\tau}{\tau}}\transpose
				%		
				%\samplespatiocormat\timenot{\timeind}%\timetotimenot{1}{\timeind}
				%		=\sum_{\tau=1}^{\timeind}\residualspatvec\timenot{\tau}
				%		\residualspatvec\transpose\timenot{\tau}
				%\expectednb[(\residualspatvec\timenot{\timeind}-\residualspatmean
				% 
				%)(\residualspatvec\timenot{\timeind}-\residualspatmean)\transpose]
				% 
				+\forgetingfactorspatio^\timeind\identitymat
				\label{eq:resspatcor}\\
				\statenoisecormat\timenot{\timeind}%\timetotimenot{1}{\timeind}
				=&\sum_{\tau=1}^{\timeind}\forgetingfactorstatenoise^{\timeind-\tau}\residualstateervec\timegiventimenot{\tau}{\tau}
				{\residualstateervec\timegiventimenot{\tau}{\tau}}\transpose
				%		
				%\samplespatiocormat\timenot{\timeind}%\timetotimenot{1}{\timeind}
				%		=\sum_{\tau=1}^{\timeind}\residualspatvec\timenot{\tau}
				%		\residualspatvec\transpose\timenot{\tau}
				%\expectednb[(\residualspatvec\timenot{\timeind}-\residualspatmean
				% 
				%)(\residualspatvec\timenot{\timeind}-\residualspatmean)\transpose]
				% 
				+\forgetingfactorstatenoise^\timeind\identitymat
				\label{eq:resstatecor}
				%\\
				%			\acom{or}
				%			
				%\spatiocormat\timenot{\timeind}%\timetotimenot{1}{\timeind}
				%			
				%=&\frac{1}{\timeind}\sum_{\tau=1}^{\timeind}\forgetingfactorspatio^{\timeind-\tau}\estsignalspatiocomp\timegiventimenot{\tau}{\tau}
				%			
				%\estsignalspatiocomp\transpose\timegiventimenot{\tau}{\tau}
				%%		
				%%\samplespatiocormat\timenot{\timeind}%\timetotimenot{1}{\timeind}
				%%		=\sum_{\tau=1}^{\timeind}\residualspatvec\timenot{\tau}
				%%		\residualspatvec\transpose\timenot{\tau}
				%		
				%%%\expectednb[(\residualspatvec\timenot{\timeind}-\residualspatmean
				%% 
				%		
				%%%)(\residualspatvec\timenot{\timeind}-\residualspatmean)\transpose]
				%% 
				%		+\frac{1}{\timeind}\identitymat
				%		\label{eq:resspatcor1}\\
				%		
				%\statenoisecormat\timenot{\timeind}%\timetotimenot{1}{\timeind}
				%		
				%=&\frac{1}{\timeind}\sum_{\tau=1}^{\timeind}\forgetingfactorstatenoise^{\timeind-\tau}
				% 
				%		\residualstateervec\timegiventimenot{\tau}{\tau}
				%		
				%\residualstateervec\transpose\timegiventimenot{\tau}{\tau}
				%		
				%%%\expectednb[(\residualspatvec\timenot{\timeind}-\residualspatmean
				%% 
				%		
				%%%)(\residualspatvec\timenot{\timeind}-\residualspatmean)\transpose]
				%% 
				%		+\frac{1}{\timeind}\identitymat
				%		\label{eq:resstatecor1}
				\end{align}
			\end{subequations}
			instead of 
			$\spatiocorsmat\timenot{\timeind}$ 
			and $\statenoisecorsmat\timenot{\timeind}$,
			where $\forgetingfactorstatenoise, \forgetingfactorspatio\in(0,1)$  are forgetting factors
			that  weigh exponentially 
			past 
			observations, and ensure 
			invertibility of matrices 
			$\spatiocormat\timenot{\timeind}$ and $
			\statenoisecormat\timenot{\timeind}$.
			%; see e.g. 
			%~\cite{angelosante2010online}. 
			%	\acom{3 choises for forgetting factors 
			%	write as a function of the inputs}\acom{comment about the 
			%divergence in 
			%	nonconvex local 
			%	minima }
			%\begin{subequations}
			%	\begin{align}
			%	\label{eq:spatiomultkerwithcor}
			%%	
			%%\kernelcoefspatiovecest\timenot{\timeind}&=\underset{\kernelcoefspatiovec\ge
			%%	 \bm 0}{\argmin} 
			%%\Tr{(\samplespatiocormat\timenot{\timeind}\genkernelmat\inv
			%%	 \kernelcoefnot{\kernelcoefspatiovec})}
			%%	+\kernelcoefspatioreg\|\kernelcoefspatiovec\|_2^2\\
			%	
			%\kernelcoefspatiovecest\timenot{\timeind}&=\underset{\kernelcoefspatiovec\ge
			%		\bm 0}{\argmin} 
			%\Tr{(\spatiocormat\timenot{\timeind}\fullkernelspatiomat\inv
			%		\kernelcoefnot{\kernelcoefspatiovec})}
			%	+\kernelcoefspatioreg\|\kernelcoefspatiovec\|_2^2\\
			%	
			%\kernelcoefstatenoisevecest\timenot{\timeind}&=\underset{\kernelcoefstatenoisevec\ge
			%	 \bm 
			%		0}{\argmin}
			%	
			%\Tr{(\statenoisecormat\timenot{\timeind}\fullkernelstatenoisemat\inv
			% 
			%	\kernelcoefnot{\kernelcoefstatenoisevec})}
			%	+\kernelcoefstatenoisereg\|\kernelcoefstatenoisevec\|_2^2
			%	\label{eq:stateermultkerwithcor}
			%	\end{align}
			%\end{subequations}
			\myitem\cmt{online update}Moreover, 
			$\spatiocormat\timenot{\timeind}$ and $
			\statenoisecormat\timenot{\timeind}$ can be updated recursively as
			\begin{subequations}
				\label{eq:rescorrec}
				\begin{align}
				\spatiocormat\timenot{\timeind}%\timetotimenot{1}{\timeind}
				=&\forgetingfactorspatio\spatiocormat\timenot{\timeind-1}+
				\estsignalspatiocomp\timegiventimenot{\timeind}{\timeind}
				{\estsignalspatiocomp\timegiventimenot{\timeind}{\timeind}}\transpose
				%		
				%\samplespatiocormat\timenot{\timeind}%\timetotimenot{1}{\timeind}
				%		=\sum_{\tau=1}^{\timeind}\residualspatvec\timenot{\tau}
				%		\residualspatvec\transpose\timenot{\tau}
				%\expectednb[(\residualspatvec\timenot{\timeind}-\residualspatmean
				% 
				%)(\residualspatvec\timenot{\timeind}-\residualspatmean)\transpose]
				% 
				\label{eq:resspatcorsec}\\
				%		
				%\spatiocormat\timenot{\timeind+1}%\timetotimenot{1}{\timeind}
				%		
				%=&\frac{\timeind\forgetingfactorspatio}{\timeind+1}\spatiocormat\timenot{\timeind}+
				%		
				%\estsignalspatiocomp\timegiventimenot{\timeind+1}{\timeind+1}
				%		
				%\estsignalspatiocomp\transpose\timegiventimenot{\timeind+1}{\timeind+1}
				%		+\frac{1-\forgetingfactorspatio}{\timeind+1}\identitymat
				%		%		
				%%\samplespatiocormat\timenot{\timeind}%\timetotimenot{1}{\timeind}
				%		%		
				%%=\sum_{\tau=1}^{\timeind}\residualspatvec\timenot{\tau}
				%		%		\residualspatvec\transpose\timenot{\tau}
				%		
				%%%\expectednb[(\residualspatvec\timenot{\timeind}-\residualspatmean
				%% 
				%		
				%%%)(\residualspatvec\timenot{\timeind}-\residualspatmean)\transpose]
				%% 
				%		\label{eq:resspatcorsec}\\
				\statenoisecormat\timenot{\timeind}%\timetotimenot{1}{\timeind}
				=&\forgetingfactorstatenoise\statenoisecormat\timenot{\timeind-1}+
				\residualstateervec\timegiventimenot{\timeind}{\timeind}
				{\residualstateervec\timegiventimenot{\timeind}{\timeind}}\transpose
				\label{eq:resstatecorsec}
				\end{align}
			\end{subequations}
			which significantly reduces the required memory for the computation 
			with respect 
			to~\eqref{eq:rescor}, since $\{	
			\estsignalspatiocomp\timegiventimenot{\tau}{\tau}
			,\residualstateervec\timegiventimenot{\tau}{\tau}\}_{\tau=1}^{\timeind-1}$ need not be stored.
			%\acom{requirements for the expansion coefficients
			%	update not dependent on the amount of historical data
			%	available.}\acom{explain?}
		\end{myitemize}
		
	\end{myremarkhere}

\section{Simulations}
\label{sec:sims}

\begin{myitemize}
	\myitem \cmt{traffic dataset loukas}
	\myitem \cmt{brain dataset}
	\myitem \cmt{economic dataset http://cow.la.psu.edu/COW2}%20Data/Trade/Trade.html}
\end{myitemize}

This section evaluates the performance of the developed algorithms by means of 
numerical tests with 
synthetic and real data. The proposed algorithms are compared with:
\begin{myitemize}
	\myitem\cmt{LMS}(i) The least mean-square (LMS) algorithm 
	in~\cite{lorenzo2016lms} with step 
	size 
	$\lmsstepsize$; and
	\myitem\cmt{DLSR}(ii) the distributed least-squares reconstruction
	(DLSR) algorithm~\cite{wang2015distributed} with step sizes $\dlsrstepsize$ 
	and  $\dlsrbeta$. Both 
	LMS and DLSR can track slowly time-varying  $\bandwidth$-bandlimited graph 
	signals. 

	\end{myitemize}\cmt{NMSE definition}The performance of the aforementioned approaches is 
quantified through the 
normalized 
mean-square error (NMSE)
\begin{align*}
\text{NMSE}%(\{\sampleset\timenot{\tau}\}_{\tau=1}^\timeind)
:=\frac{\expectednb\big[\sum_{\tau=1}^{\timeind}
	\|\samplemat^c\timenot{\tau}(\signalvec\timenot{\tau}
	-\signalestvec\timenot{\tau|\tau})\|^2_2\big]}
{\expectednb\big[\sum_{\tau=1}^{\timeind}	 
	\|\samplemat^c\timenot{\tau}\signalvec
	\timenot{\tau}\|^2_2\big]}
\end{align*}
where the expectation is taken over the sample locations,  and  
$\samplemat^c\timenot{\tau}$ is an
$(\vertexnum\!-\!\samplenum\timenot{\tau})\!\times\!\vertexnum$ matrix
comprising the rows of $\identitymat_\vertexnum$ whose indices are
not in $\sampleset\timenot{\timeind}$. For  all tests,
$\sampleset\timenot{\timeind}$ is chosen uniformly at random without 
replacement over $\vertexset$,
and kept 
constant over time; that is,
$\sampleset\timenot{\timeind}=\sampleset,~\forall\timeind$. 
The parameters of different algorithms were selected using cross-validation to 
minimize their NMSE. Notice 
that our MKriKF, which learns the kernel that ``best" fits the data, 
requires minimal parameter tuning.

\subsection{Numerical tests on synthetic data}
\begin{myitemize}
	\myitem\cmt{Synthetic dataset}
%The synthetic experiments will evaluate the 
%perfomance of the KeKriKF 
%	algorithm. 
%		\myitem\cmt{KeKriKF synth}
%		Specifically, algorithm~\ref{algo:krkalmanfilter} admits the following 
%		configuration:
%		\begin{myitemize}
%			\myitem\cmt{kernel spatio}$\fullkernelspatiomat\timenot{\timeind}$ is a diffusion 
%			kernel 	with 
%			parameter  $\sigma$ in the first experiment and the bandlimited kernel with parameters 
%			$\beta$,   
%			$\bandwidth$ (cf. 
%			Table~\ref{tab:spectralweightfuns}) in the second experiment;
%			\myitem\cmt{kernel trans noise}
%			$\fullkernelstatenoisemat\timenot{\timeind}=\plantnoiseweight\identitymat_\vertexnum$
%			with parameter 
%			$\plantnoiseweight>0$;
%			\myitem\cmt{reg par}regularization parameters $\regparone=1$, $\regpartwo=1$;
%			\myitem\cmt{trans matrix}and a transition matrix 
%			$\adjtransgraphmat\timetimenot{\timeind}{\timeind-1}=\transweight 
%			(\adjacencymat\timenot{\timeind-1}+\identitymat_\vertexnum)$
%			with parameter $\transweight>0$.
%		\end{myitemize}
%	
	\begin{myitemize}\myitem\cmt{evaluate decomposition}\begin{myitemize}\myitem\cmt{Graph 
	description}To construct a graph, consider the dataset in~\cite{collegeMSGSnap}, which contains 
	timestamped	messages among
		students at the University 
		of California, Irvine, exchanged over a social network during 90 
		days.
		 The 
		sampling interval $\timeind$ is one day.  A graph is constructed such that the edge weight
		$\adjacencymatentry 
		\timevertexvertexnot{\timeind}{\vertexind}{\vertexindp}
		$ 
		counts the 	number of messages exchanged between student $\vertexind$ 
		and $\vertexindp$ in 
		the $k$-th month, where $k=1,2,3$ and $30(k-1)+1\le\timeind\le30k$.
		Hence, $\adjacencymat\timenot{\timeind}$ changes across months. 
		%A 
		%network was created where
		%$\{\adjacencymatentry\timevertexvertexnot{\timeind}{\vertexind}{\vertexindp}
		%\}_{\timeind=30(k-1)+1}		 
		%^{\timeind=30k}$ 
		%counts the 	number of 	messages exchanged between student 
		%$\vertexind$ and $\vertexindp$ in 
		%the $k$-th month. 
		%The resulting topology changes across months.
		 A subset of 
		$\vertexnum=310$ users  for which 
		$\adjacencymat\timenot{\timeind}$ corresponds to a connected graph 
		$\forall \timeind$ is 
		selected.
		\myitem\cmt{Signal description}At each  $\timeind$, $\truesignal\timenot{\timeind}$ was 
		generated 
		by superimposing a $\bandwidth$-bandlimited graph function with 
		$\bandwidth=5$ 
		and 
		a spatio-temporally correlated signal. Specifically,
		$\truesignal\timenot{\timeind}=\truesignalspatiocomp
		\timenot{\timeind} +\truesignalspatiotempcomp\timenot{\timeind} 
		=\sum_{i=1}^{5} 
		\gamma^{i}\timenot{\timeind} 
		\bm u^{i} 
		\timenot{\timeind}
		+\truesignalspatiotempcomp\timenot{\timeind}$, where 
		$\{\gamma^{i}\timenot{\timeind}\}_{i=1}^{5}~\sim\mathcal{N}(0,1)$ for 
		all $\timeind$, while $\{\bm 
		u^{i}\timenot{\timeind}\}_{i=1}^{5}$ denote the 
		eigenvectors
		associated with the 5 smallest eigenvalues of 
		$\laplacianmat\timenot{\timeind}$, and
		$\truesignalspatiotempcomp\timenot{\timeind}$ is generated according to 
		\eqref{eq:transmod} with
		$\adjtransgraphmat\timetimenot{\timeind}{\timeind-1}= 
		0.03(\adjacencymat\timenot{\timeind-1}+\identitymat_\vertexnum)$, 
		$\plantnoisevec\sim\mathcal{N}(\bm{0},\bm{C}_\eta)$, and $\bm{C}_\eta$ is a diffusion kernel 
		with $\sigma=0.5$.
		Function $\signalfun\timevertexnot{\timeind}{\vertexind}$ is therefore 
		smooth with respect to the graph and can be interpreted e.g. as the 
		time that the  
		$\vertexind$-th student spends on  the specific social network during the $\timeind$-th day. 
	
		\myitem\cmt{Experiment 1:NMSE vs time: decomposition
			\ra Fig.~\ref{fig:nmseDecomp}}The first experiment justifies the  
		proposed decomposition by assessing the impact of 
		dropping either $\truesignalspatiocomp\timenot{\timeind}$ or 
		$\truesignalspatiotempcomp\timenot{\timeind}$ from the right hand side 
		of 
		\eqref{eq:decomp}.
				\myitem\cmt{parameters}The 
				KriKF algorithm uses diffusion 
				kernels  	$\fullkernelspatiomat\timenot{\timeind}$ and 
				$\fullkernelstatenoisemat\timenot{\timeind}$	with parameters  $\sigma=1.5$  and 
				$\sigma=0.5$, respectively.
		\begin{myitemize}\myitem\cmt{explain fig}
			Fig.~\ref{fig:nmseDecomp} depicts the NMSE with $\samplenum=217$ 
		for the
			KeKriKF; 
			\myitem\cmt{KF}the Kalman filter (KF) 
			estimator, which results from setting 
			$\truesignalspatiocomp\timenot{\timeind}=\bm 0$ for all 
			$\timeind$ in the KeKriKF;
			\myitem\cmt{KR} as well 
			as kernel Kriging (KKr), which the KeKriKF reduces to if 
			$\truesignalspatiotempcomp\timenot{\timeind}=\bm 0$ for all 
			$\timeind$.
			\myitem\cmt{Punchline} As observed, KeKriKF, which accounts for both summands 
			in~\eqref{eq:decomp}, 
			outperforms those algorithms 
			that account for only one of them.
			Moreover, the low NMSE of KeKriKF in reconstructing the 
			$N-S=310-217=93$ unavailable node values reveals that this 
			algorithm is capable of 
			efficiently 
			capturing the spatial as well as the temporal dynamics over time-varying topologies.
				\begin{figure}[t]
					%\centering{\input{1202.tex}}
					\centering{\includegraphics[width=\linewidth]
						{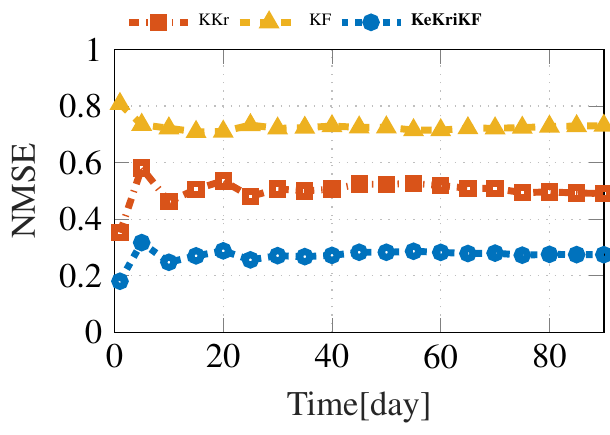}}%
					%{\includegraphics[width=0.47\linewidth]{misc/KFonGSimulations_3219-1}}%\vspace{-1em}
					\hfill
					\caption{NMSE of function estimates 
					($\regparone=\regpartwo=1$).} % temperature 
					%estimates. 
					%	($\regparone=1$, 
					%		$\regpartwo=1$, $\dlsrstepsize
					%		=1.6$, $\dlsrbeta=0.5$, $\lmsstepsize =0.6$, 
					%		$\transweight=10^{-3}$, 
					%		$\gausmeanstatenoise=10^{-5}$, $\gausstdstatenoise=10^{-6}$, 
					%		$\gausmeanspatio=2$, 
					%		$\gausstdspatio=0.5$, $\rkhsspationum=40$, $\rkhsstatenoisenum=40$)}
					%\label{fig:recon}
					\label{fig:nmseDecomp}
					%\vspace{-1em}
				\end{figure}
%			\begin{figure}[t]
%				\centering
%				%	    \hspace{-0.3cm}
%				\includegraphics[width=8.5cm]{KrKFonGSimulations_1202-1}
%				\caption{NMSE of function estimates. ($\sigma=1.5$, 
%					$\transweight=0.028$, 
%					$\plantnoiseweight=0.05$)}
%				\label{fig:nmseDecomp}
%				%\vspace{-2em}
%			\end{figure}	
		\end{myitemize}
	\end{myitemize} 
	%\vspace{-1em}
	
	\myitem\cmt{evaluate robustness}Next, the robustness of KeKriKF is 
	evaluated 
   when the connectivity of $\graph\timenot{\timeind}$, captured by
	$\adjacencymat\timenot{\timeind}$,  exhibits abrupt 
	changes over $\timeind$. 
	\begin{myitemize}
		\myitem\cmt{Graph description}Synthetic time-varying networks of size 
		$\vertexnum=81$ 
		were generated using the Kronecker product model, which effectively captures properties 
		of 
		real graphs~\cite{leskovec2010kronecker}. The 
		prescribed ``seed matrix" 
			\begin{align*}
			\seedzeromat\define\left[ 
			\begin{array}{ccc}
			1  & 0.1 & 0.7\\
			0.3  & 0.1 & 0.5 \\
			0  & 1 & 0.1
			\end{array}
			\right]
			\end{align*}
		produces the  $\vertexnum\times\vertexnum$ matrix 
		$\seedmat\define\seedzeromat\otimes\seedzeromat\otimes\seedzeromat\otimes\seedzeromat$, 
		where $\otimes$ denotes the Kronecker product.
		An initial adjacency matrix 
		$\adjacencymat\timenot{0}$ was constructed with entries 
		$\adjacencymatentry\timevertexvertexnot{0}{\vertexind}{\vertexindp}~\forall\vertexind$,
		$\adjacencymatentry\timevertexvertexnot{0}{\vertexind}{\vertexindp}\sim\text{Bernoulli}
		(\seedmatentry\vertexvertexnot{\vertexind}{\vertexindp})$ for $
		\vertexind>\vertexindp$, and 
		$\adjacencymatentry\timevertexvertexnot{0}{\vertexind}{\vertexindp}= 
		\adjacencymatentry\timevertexvertexnot{0}{\vertexindp} {\vertexind}$ 
		for $\vertexind 
		<\vertexindp$.
		Next, the following time-varying graph model was generated: at each 
		$\timeperiodch=10\kappa,~\kappa=1,2,\ldots$, each entry of	
		$\adjacencymat\timenot{\timeperiodch}$ changes with 
		probability 
		{$p\vertexvertexnot{\vertexind}{\vertexindp}={\sum_{k}\adjacencymatentry
		\timevertexvertexnot{\timeperiodch}{\vertexind}{k}\sum_{l}\adjacencymatentry
		\timevertexvertexnot{\timeperiodch}{l}{\vertexindp}}/
		{\sum_{k}\sum_{l}\adjacencymatentry
		\timevertexvertexnot{\timeperiodch}{k}{l}}$} as
		$
	\adjacencymatentry\timevertexvertexnot{\timeperiodch+1}{\vertexind}{\vertexindp}=
	\adjacencymatentry\timevertexvertexnot 
	{\timeperiodch}{\vertexind}{\vertexindp}+ 
	|\adjentrynoise\timevertexvertexnot 
	{\timeperiodch}{\vertexind}{\vertexindp}|$ 
	 for $\vertexind>\vertexindp$
	where
		$\adjentrynoise\timevertexvertexnot 
		{\timeperiodch}{\vertexind}{\vertexindp} 
		\sim\mathcal{N}(0,\adjentrynoisevar)$ and $		
		\adjacencymatentry\timevertexvertexnot 
		{\timeperiodch+1}{\vertexindp}{\vertexind}=		
		\adjacencymatentry\timevertexvertexnot{\timeperiodch+1}{\vertexind}
		{\vertexindp}$ 
		for $\vertexind<\vertexindp$. This
		choice of $p\vertexvertexnot{\vertexind}{\vertexindp} $ is based on the ``rich get richer" 
		attribute
		of real networks, where new connections 
		are formed between nodes with high degree~\cite{leskovec2010kronecker}. Moreover, the edge 
		$(v_\vertexind,v_\vertexindp)$ is deleted at each $\timeperioddel=20\kappa,~\kappa=1,2,\ldots$ 
		with probability $0.1$;	that is,
		$\adjacencymatentry 
		\timevertexvertexnot{\timeperioddel+1}{\vertexindp}{\vertexind}=		
		\adjacencymatentry\timevertexvertexnot{\timeperioddel+1}{\vertexind}{\vertexindp}=0$, as long 
		as the 
		graph remains connected. 
		%The parameter $\adjentrynoisevar$ determines 
	%the magnitude of the edge weight change, and 
		By varying 
		$\adjentrynoisevar$, we obtain  
		 different time-varying graphs.
%		\acom{explain how to create adj from seed}
%		\acom{muliple time-varying models of graphs}
		\myitem\cmt{Signal description}A graph function was generated for each 
		time-varying graph   as follows\begin{align}
		\rkhsvec\timenot\timeind=\forgetingfactorsynth\adjacencymat\timenot{\timeind}\rkhsvec\timenot
		{\timeind-1}+\sum_{i=1}^{10}\gamma^{(i)}\timenot{\timeind} 
		\bm u^{(i)} 
		\timenot{\timeind}
		\end{align} 
		where $\forgetingfactorsynth=10^{-2}$ is a forgetting factor,
		$\sum_{i=1}^{10}\gamma^{(i)}\timenot{\timeind} 
		\bm u^{(i)} 
		\timenot{\timeind}$ is a graph-bandlimited component with 
		$\gamma^{(i)}\timenot{\timeind}\sim\mathcal{N}(0,1)$,
		and  $\{\bm 
		u^{(i)}\timenot{\timeind}\}_{i=1}^{10}$ are the 
		eigenvectors
		associated with the 10 smallest eigenvalues of $\laplacianmat\timenot{\timeind}$. 
		\myitem\cmt{parameters}Algorithm~\ref{algo:krkalmanfilter} employs a bandlimited kernel with 
		$\beta=10^{3}$ and 
		$\bandwidth$ for  $\fullkernelspatiomat\timenot{\timeind}$, a diffusion 
					kernel with $\sigma=0.5$  for
					$\fullkernelstatenoisemat\timenot{\timeind}$, and
					$\adjtransgraphmat\timetimenot{\timeind}{\timeind-1}= 
					10^{-3}(\adjacencymat\timenot{\timeind-1}+\identitymat_\vertexnum)$.
		\myitem\cmt{Experiment 2:NMSE vs time: robustness
			\ra Fig.~\ref{fig:nmsedifmodel}}
		\begin{myitemize}\myitem\cmt{explain fig}
		Fig.~\ref{fig:nmsedifmodel} plots the NMSE  of the KeKriKF algorithm  as a function of 
		$\adjentrynoisevar$, which determines how rapidly the graph changes.
			\myitem\cmt{Punchline}As observed, the KeKriKF algorithm can effectively cope with different 
			degrees of time variation.
							\begin{figure}[t]
								%\centering{\input{1006.tex}}
								\centering{\includegraphics[width=\linewidth]
									{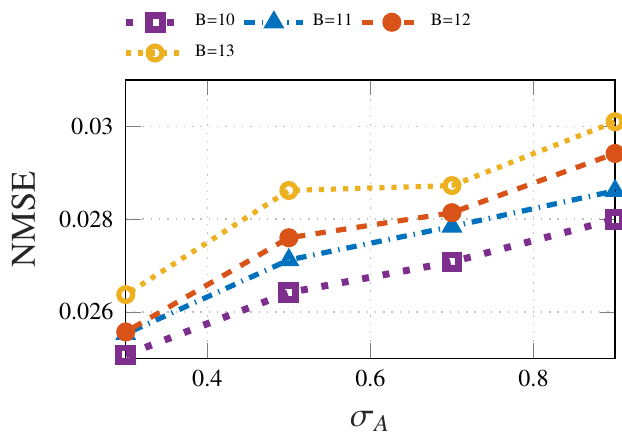}}%
								%{\includegraphics[width=0.47\linewidth]{misc/KFonGSimulations_3219-1}}%\vspace{-1em}
								\hfill
								\caption{NMSE of KeKriKF for different 
								time-varying graphs ($\samplenum=65$, 
								$\regparone=\regpartwo=1$).} 
								\label{fig:nmsedifmodel}
								%\vspace{-1em}
							\end{figure}
		\end{myitemize}
	\end{myitemize}
	\end{myitemize}
	\end{myitemize}
	\subsection{Temperature prediction}
		\begin{myitemize}

	\myitem\cmt{temperature dataset}
	\begin{myitemize}
		\myitem\cmt{Description}Consider the dataset \cite{USATemp} provided by the  National Climatic 
		Data 
		Center, which 
		comprises  
		hourly temperature measurements at $\vertexnum=109$ measuring stations 
		across the continental 
		United States in 2010.
		%			\acom{rephrase}  Temperature reconstruction 
		%			over a network of stations has been heavily 
		%			used 
		%			to 	evaluate the performance of graph inference algorithms 
		%			\cite{romero2016spacetimekernel,wang2015distributed}.
		\begin{myitemize}\myitem\cmt{adjacency}A time-invariant graph was constructed  as in 
			\cite{romero2016spacetimekernel}, based on geographical distances.
			%			using 
			%			\begin{align}
			%			\spaceadjacencymatentry\vertexvertexnot
			%			{\vertexind}{\vertexindp}=\frac{\exp{\{-d_{\vertexind,\vertexindp}^2\}}}
			%			{\sqrt{\sum_{j\in{\mathcal{N}_\vertexind^k}}\exp{\{-d_{\vertexind,j}^2\}}
			%					\sum_{l\in{\mathcal{N}_\vertexindp^k}}\exp{\{-d_{\vertexindp,l}^2\}}}}
			%			\end{align}
			%			where $\mathcal{N}_\vertexindp^k$ is the set comprising the $k=7$ 
			%			nearest neighbors of 
			%			station $\vertexindp$, and $d_{\vertexind,\vertexindp}$ is the 
			%			geographical 
			%			distance between stations $\vertexind$ and $\vertexindp$. For this 
			%			tests,
			%			the neighborhood is defined based on $d_{\vertexind,\vertexindp}$, 
			%			which is justified since geographically close stations tend to measure
			%			similar temperature values.
			\myitem\cmt{signal} The value
			$\signalfun\timevertexnot
			{\timeind}{\vertexind}$ represents the $\timeind$-th
			temperature sample recorded at the $\vertexind$-th  station. The 
			sampling interval is one hour for the first 
			experiment, and one day for the second.
		\myitem\cmt{configur algo} 
			\begin{myitemize}
				\myitem\cmt{kekrikf}KeKriKF
					\myitem\cmt{kernel spatio}employs diffusion 
					kernels with 
					parameter  $\sigma=1.8$ for $\fullkernelspatiomat\timenot{\timeind}$,
					\myitem\cmt{kernel trans noise}
					$\fullkernelstatenoisemat\timenot{\timeind}=10^{-5}\identitymat_\vertexnum$,
					\myitem\cmt{trans matrix}and a transition matrix 
					$\adjtransgraphmat\timetimenot{\timeind}{\timeind-1}=5\cdot10^{-4} 
					(\adjacencymat\timenot{\timeind-1}+\identitymat_\vertexnum)$.
				\myitem\cmt{mkkrkf}MKriKF is 
				configured as 
				follows:
				    $\fullkernelspatiomatdict$ contains 
					$\rkhsspationum=40$ 
					diffusion kernels with parameters 
					$\{\sigma\kernelindnot{\rkhsind}\}_{\rkhsind=1}^{40}$ with $\sigma 
					\kernelindnot{\rkhsind}\sim\mathcal{N}(2,0.5),\forall\rkhsind$;
				 $\fullkernelstatenoisematdict$ contains
					$44$ 
					diffusion kernels with parameters 
					$\{\sigma\kernelindnot{\rkhsind}\}_{\rkhsind=1}^{44}$, where $\sigma 
					\kernelindnot{\rkhsind}\sim\mathcal{N}(1,0.2),\forall\rkhsind$, and an
					identity kernel 				
					$\fullkernelstatenoisemat\kernelindnot{45}=\identitymat_\vertexnum$.
					%			;
					%			\myitem\cmt{trans matrix}and a transition matrix 
					%			$\adjtransgraphmat\timetimenot{\timeind}{\timeind-1}=\transweight 
					%			(\adjacencymat\timenot{\timeind-1}+\identitymat_\vertexnum)$
					%			with parameter $\transweight$.
				
			\end{myitemize}
			
		\end{myitemize}\myitem\cmt{Experiment 3:temperature vs time
		\ra Fig.~\ref{fig:track}}
	%Next, the performance  of the 
	%different reconstruction 
	%algorithms is 
	%evaluated in tracking the temperature values.
	\begin{myitemize}\myitem\cmt{explain fig}Fig.~\ref{fig:track} depicts the true temperature  
		along 
		with its estimates for a station $\vertexind$ that is not 
		sampled, meaning $\vertexind\notin\sampleset$, with $\samplenum=44$.
		\myitem\cmt{result}Clearly, KeKriKF accurately 
		tracks 	the 
		temperature by exploiting spatial and temporal dynamics, but MKriKF  outperforms KeKriKF by 
		learning those dynamics from the data.
%		On 
%		the 
%		other hand, DLSR and 
%		LMS cannot capture the fast signal variations. 
		The random sampling set selection heavily affects 
		performance of the LMS algorithm; for adaptive selection of $\sampleset$ 
		see~\cite{lorenzo2016lms}.  
							\begin{figure}[t]
								%\centering{\input{25282.tex}}
								\centering{\includegraphics[width=\linewidth]
									{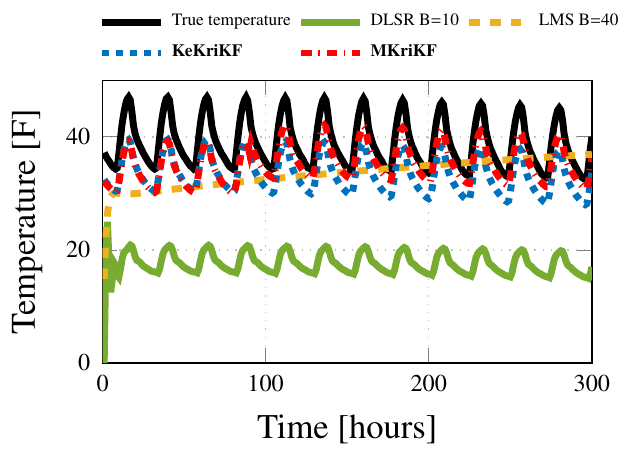}}%
								%{\includegraphics[width=0.47\linewidth]{misc/KFonGSimulations_3219-1}}%\vspace{-1em}
								\hfill
								\caption{True and estimated temperature values 
									($\bandwidth=5$, $\dlsrstepsize
									=1.2$, $\dlsrbeta=0.5$, $\lmsstepsize 
									=1.5$, $\regparone=\regpartwo=1$).}
								\label{fig:track}
								%\vspace{-1em}
							\end{figure}
							
%		\begin{figure}[t]
%			\centering
%			
%\includegraphics[width=8.5cm]{KrKFonGSimulations_25282-1}%\vspace{-1em}
%			\caption{True temperature values along with the 
%				estimated ones. 
%				($\bandwidth=5$, $\dlsrstepsize
%				=1.2$, $\dlsrbeta=0.5$, $\lmsstepsize =1.5$)}
%			\label{fig:track}
%			%\vspace{-1em}
%		\end{figure}
	\end{myitemize}
	
	\myitem\cmt{Experiment 4:NMSE vs time
	\ra Fig.~\ref{fig:recon}}Fig.~\ref{fig:recon} 
compares 
the NMSE 
of all considered
approaches for 
$\samplenum=44$. Observe the superior performance of the proposed reconstruction methods, which 
in this scenario exhibit roughly the same NMSE.

\begin{figure}
			%\centering{\input{25292.tex}}
	\centering{\includegraphics[width=\linewidth]
		{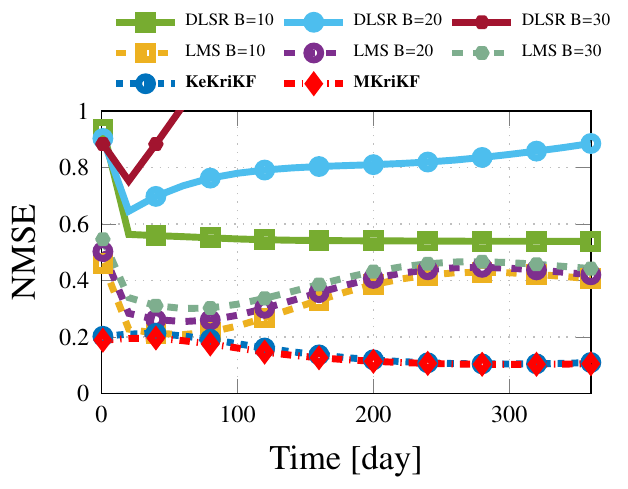}}%
	\caption{NMSE of temperature estimates ($\dlsrstepsize
		=1.6$, $\dlsrbeta=0.5$, $\lmsstepsize =1.5$, 
	 $\kernelcoefspatioreg=10^{5}$).}
	\label{fig:recon}
	%\vspace{-1em}
\end{figure}

\end{myitemize}
\subsection{GDP prediction}
\myitem\cmt{GDP dataset}
\begin{myitemize}
		\myitem\cmt{Description}
		The  next dataset is provided by the World Bank  
		Group~\cite{wordbank},
		and  comprises 
		gross domestic product (GDP) per capita  for $\vertexnum=127$ countries  for the years 
		1960-2016. 
		%amounts to the monetary measure of the market value of all goods and services 
		%The second dataset is 
		%~\cite{USATemp}, and 
		%comprises  
		%hourly temperature measuments at $\vertexnum=109$ measuring stations across the continental 
		%\textsl{}United States in 2010.
		%			\acom{rephrase}  Temperature reconstruction 
		%			over a network of stations has been heavily 
		%			used 
		%			to 	evaluate the performance of graph inference algorithms 
		%			\cite{romero2016spacetimekernel,wang2015distributed}.
		\begin{myitemize}\myitem\cmt{adjacency}
		A time-invariant graph was constructed using the correlation between the GDP 	of different 
		countries  for the first 
		25 years.
			\myitem\cmt{signal}The graph function
			$\signalfun\timevertexnot
			{\timeind}{\vertexind}$ denotes the
			GDP  reported at the $\vertexind$-th  country and
			$\timeind$-th year for $\timeind=1985,\ldots,2016$. 
			\end{myitemize}
			\myitem\cmt{kernels}
				The graph Fourier transform of the GDP in the first 25 years defined as 		
				$\fouriersignalfun\vertexnot{\vertexind}\define\laplacianevec_\vertexind 
						\transpose
					\signalvec~\forall\vertexind$, where $\laplacianevec_\vertexind$ denotes the $n$-th 
					eigenvector of the Laplacian matrix; see~\cite{shuman2013emerging}, shows 
				that the graph 
				frequencies
				$\fouriersignalfun_k$ take
				small values for $4<k<123$, and large values otherwise.
				Motivated by the aforementioned observation, the KeKriKF is 
				configured with a band-reject 
				kernel
				$\fullkernelspatiomat$
				% that 
				%results after applying $\frequencyweightfun(\lambda_\vertexind)=\beta$ for 
				%$k\leq\vertexind\leq\vertexnum- l$ and $\frequencyweightfun(\lambda_\vertexind)=1/\beta$ 
				%otherwise 
				%n~\eqref{eq:laplaciankerneldef}
				 with $k=6, l=6, \beta=15$; see Table~\ref{tab:spectralweightfuns},  
				$\fullkernelstatenoisemat=10^{-3}\identitymat_\vertexnum$, and 
				$\adjtransgraphmat\timetimenot{\timeind}{\timeind-1}=10^{-5}
				(\adjacencymat\timenot{\timeind-1}+\identitymat_\vertexnum)$. MKriKF adopts 
				a $\fullkernelspatiomatdict$ with $\rkhsspationum=16$ 
				band-reject kernels with $k\in[2,5],$ $ l\in[1,4],$ $ \beta=15$, and
				a $\fullkernelstatenoisematdict$ with
				$60$ 
				diffusion kernels with parameters 
				$\{\sigma\kernelindnot{\rkhsind}\}_{\rkhsind=1}^{60}$, where $\sigma 
				\kernelindnot{\rkhsind}\sim\mathcal{N}(2,0.5),\forall\rkhsind$, and an
				identity kernel
				$\fullkernelstatenoisemat\kernelindnot{61}=\identitymat_\vertexnum$.
				
			\myitem\cmt{tracking}
		    Fig.~\ref{fig:trackgdp} depicts the 
			actual GDP 
			as well as its estimates for Greece, which is not 
			contained in the sampled countries.  Clearly, both MKriKF and 
			KeKriKF, 
			track the GDP 
			evolution over the years with greater accuracy than the considered 
			alternatives. This is expected because the graph function does not 
			adhere to the graph bandlimited model assumed
			by DLSR and LMS.
			\begin{figure}

		%	\centering{\input{75182.tex}}
		\centering{\includegraphics[width=\linewidth]
			{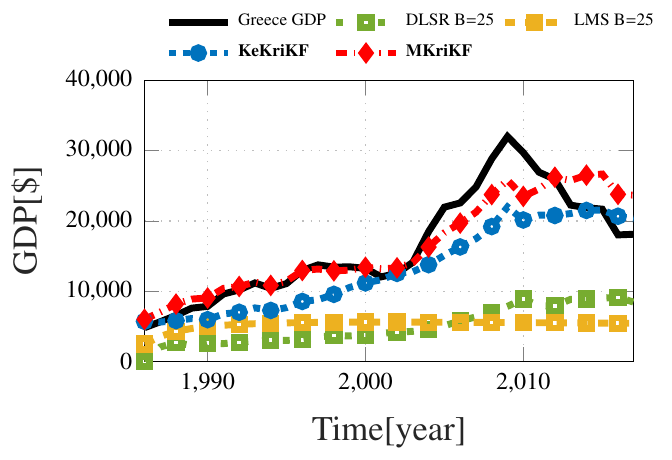}}%
			%{\includegraphics[width=0.47\linewidth]{misc/KFonGSimulations_3219-1}}%\vspace{-1em}
			\hfill
				\caption{Greece GDP values along with the estimated ones   ($\samplenum=38$, $\dlsrstepsize
					=1.6$, $\dlsrbeta=0.4$, $\lmsstepsize =1.2$,
					$\kernelcoefreg=100$).}
				\label{fig:trackgdp}
				%\vspace{-1em}
			\end{figure}
	
		\myitem\cmt{reconstruction}Fig.~\ref{fig:recongdp} reports  NMSE over time, where the 
		proposed 
		algorithms achieve the smallest NMSE.
		The data-driven MKriKF
		outperforms KeKriKF, which is configured manually.
		 
		 			\begin{figure}
		%	\centering{\input{75192.tex}}
		 \centering{\includegraphics[width=\linewidth]
		 	{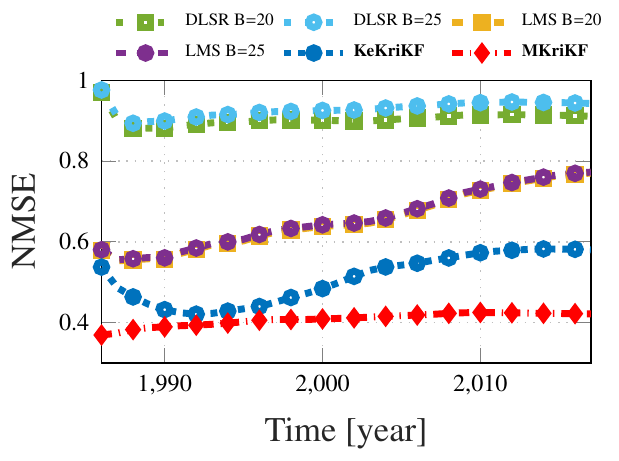}}%
		 				\caption{NMSE of GDP estimates ($\samplenum=38$, 
		 				$\dlsrstepsize
		 					=1.6$, $\dlsrbeta=0.4$, $\lmsstepsize =1.6$, 
		 					$\kernelcoefspatioreg=10^{5}$, 
		 					$\kernelcoefstatenoisereg=10^{5}$).}
		 				\label{fig:recongdp}
		 				%\vspace{-1em}
		 			\end{figure}
\end{myitemize}
\subsection{Network delay prediction}
\myitem\cmt{Network Delay dataset}
\begin{myitemize}
	\myitem\cmt{Description}
	The  last dataset records  measurements of path delays on the Internet2 
	backbone\cite{internet2}. The network comprises 9 end-nodes and $26$ directed 
	links. The 
	delays are 
	available for $\vertexnum=70$ paths at every minute. The paths connect  
	origin-destination nodes by a series of  
	links described by the path-link routing matrix  
	$\routingmat\in\{0,1\}^{\vertexnum\times26}$, 
	whose $(\vertexind,l)$ 
	entry is 
	$\routingmatentry\vertexvertexnot{\vertexind}{l}=1$ if path 
	$\vertexindp$ traverses link $l$, and 0 
	otherwise.
	\begin{myitemize}\myitem\cmt{adjacency}A  graph is constructed with each 
	vertex corresponding to 
	one of these paths, and with the time-invariant adjacency matrix
	$\adjacencymat\in\rfield^{\vertexnum\times\vertexnum}$ given by
	\begin{align}
	\label{eq:edjeweight}
		\adjacencymatentry\vertexvertexnot{\vertexind}{\vertexindp}= 
		\frac{\sum_{l=1}^{26}\routingmatentry\vertexvertexnot 
		{\vertexind}{l}  
		\routingmatentry\vertexvertexnot{\vertexindp}{l}}
		{\sum_{l=1}^{26}\routingmatentry 
		\vertexvertexnot{\vertexind}{l} + 
		\sum_{l=1}^{26} 
		\routingmatentry\vertexvertexnot{\vertexindp}{l} -
			\sum_{l=1}^{26}\routingmatentry 
			\vertexvertexnot{\vertexind}{l} 
			\routingmatentry\vertexvertexnot{\vertexindp}{l}  }
	\end{align}
	for $\vertexind,\vertexindp=1,\ldots,\vertexnum$ , $\vertexind\ne\vertexindp$. 
	Expression~\eqref{eq:edjeweight} was selected to assign a greater weight to edges connecting 
	vertices whose associated paths share a large number of links. This is 
	intuitively reasonable
	since paths with common links 
	usually experience similar delays~\cite{chua2006network}. 
		\myitem\cmt{signal}Function
		$\signalfun\timevertexnot
		{\timeind}{\vertexind}$ denotes the delay in milliseconds measured at the $\vertexind$-th  path and
		$\timeind$-th minute. 
	\end{myitemize}
	\myitem\cmt{kernels}\begin{myitemize}\myitem\cmt{kekrikf}The KeKriKF 
	algorithm
		\myitem\cmt{kernel spatio}employs a diffusion 
		kernel with 
		parameter  $\sigma=2.5$ for $\fullkernelspatiomat\timenot{\timeind}$,
		\myitem\cmt{kernel trans noise} 
		$\fullkernelstatenoisemat\timenot{\timeind}=0.002\identitymat_\vertexnum$,
		\myitem\cmt{trans matrix}and 
		$\adjtransgraphmat\timetimenot{\timeind}{\timeind-1}=0.005 
		(\adjacencymat\timenot{\timeind-1}+\identitymat_\vertexnum)$.
		\myitem\cmt{mkkrkf}The MKriKF is 
		configured as 
		follows:
		$\fullkernelspatiomatdict$ contains 
		$\rkhsspationum=40$ 
		diffusion kernels with parameters 
		$\{\sigma\kernelindnot{\rkhsind}\}_{\rkhsind=1}^{40}$ with $\sigma 
		\kernelindnot{\rkhsind}\sim\mathcal{N}(4,0.5),\forall\rkhsind$;
		$\fullkernelstatenoisematdict$ contains 
		$60$ 
		diffusion kernels with parameters 
		$\{\sigma\kernelindnot{\rkhsind}\}_{\rkhsind=1}^{60}$ with $\sigma 
		\kernelindnot{\rkhsind}\sim\mathcal{N}(1,0.1),\forall\rkhsind$, and an
		identity kernel
		$\fullkernelstatenoisemat\kernelindnot{61}=\identitymat_\vertexnum$.
		%			;
		%			\myitem\cmt{trans matrix}and a transition matrix 
		%			$\adjtransgraphmat\timetimenot{\timeind}{\timeind-1}=\transweight 
		%			(\adjacencymat\timenot{\timeind-1}+\identitymat_\vertexnum)$
		%			with parameter $\transweight$.
		
	\end{myitemize}
		\myitem\cmt{reconstruction}Fig.~\ref{fig:reconDel} depicts the NMSE 
		%of different approaches 
		when
		$\samplenum=20$. KeKriKF and MKriKF are seen to 
			outperform competing 
		methods. 
		\begin{figure}
		%	\centering{\input{85192.tex}}
		\centering{\includegraphics[width=\linewidth]
			{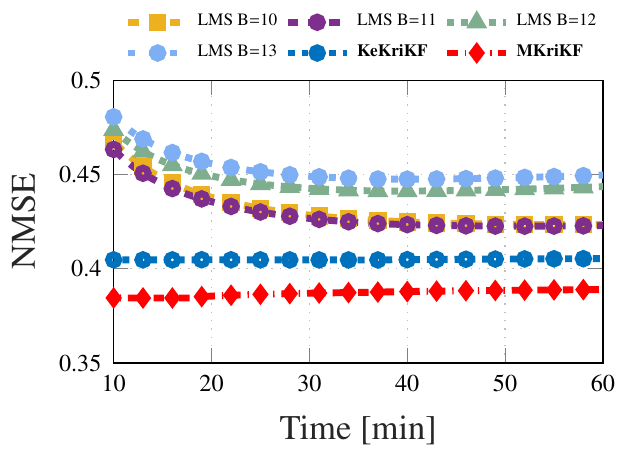}}%
			\caption{NMSE of network delay estimates ($\lmsstepsize =1.5$, 
				$\transweight=0.0005$, 
				$\kernelcoefreg=100$, $\regparone=\regpartwo=1$).}
			\label{fig:reconDel}
			%\vspace{-1em}
		\end{figure}

	\myitem\cmt{tracking}Finally, the proposed MKriKF will be evaluated in tracking the delay over the 
	network from $\samplenum=56$ randomly sampled path delays. 
	To that end, delay maps are traditionally employed, which depict the network 
	delay 
	per path over time 
	and enable  operators to perform 
	troubleshooting; see also~\cite{rajawat2014cartography}. The paths for the 
	delay maps in Fig.~\ref{fig:trackdel} 
	are sorted in  increasing 
	order  of the true delay at $\timeind=1$. Clearly, the 
	delay map recovered by  MKriKF in
	Fig.~\ref{fig:delmkrikf} visually resembles the true delay 
	map in Fig.~\ref{fig:deltrue}. 
%	 The 
%	MKL-based algorithm selects the appropriate kernel for reconstruction in a 
%	data-adaptive and 
%	online fashion, which gives MKriKF a vantage point when prior 
%	information 
%	about 
%	the target function is limited.
\begin{figure}
	\def\tabularxcolumn#1{m{#1}}
	\begin{tabularx}{\linewidth}{@{}cX@{}}
		\begin{tabular}{c}
	\subfloat[True delay]{
		\label{fig:deltrue}
	%	\hspace{0.9cm}
		%\input{true-86192.tex}
	\centering{\includegraphics[width=\linewidth]
		{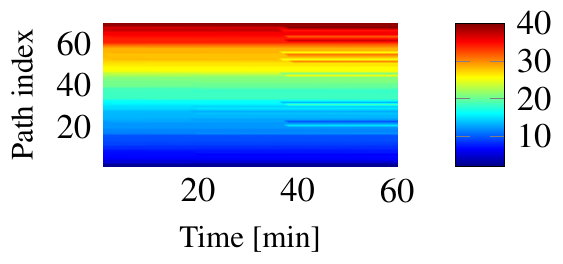}}%
	}
%	\\
%		\subfloat[LMS B=40]{
%	\label{fig:dellms}
%	\input{lms.tex}
%} 
%\\	\subfloat[KeKriKF]{
%\label{fig:delkekrikf}
%\input{kkrkfmap.tex}
%} 
\\	\subfloat[MKriKF]{
\label{fig:delmkrikf}
\hspace{-2cm}
\centering{\includegraphics[width=0.7\linewidth]
	{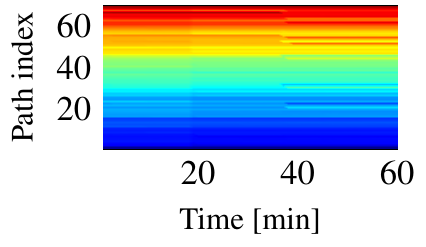}}
} 
		\end{tabular}
	\end{tabularx}
	\caption{True and estimated network delay map for $\vertexnum=70$ paths
	($\kernelcoefstatenoisereg=100$, $\regparone=\regpartwo=1$).}
	\label{fig:trackdel}
\end{figure}
\end{myitemize}

\end{myitemize}

\section{Conclusions}
%\vspace{-0.5em}
\label{sec:concl}
\begin{myitemize}
	\myitem\cmt{main idea of paper}This paper introduced online estimators to 
	reconstruct dynamic 
	functions over (possibly dynamic) graphs.
	\myitem\cmt{list contributions}In this context, the function to be estimated was decomposed in two 
	parts: one capturing the spatial dynamics, and the other jointly modeling 
	spatio-temporal dynamics by means of a state-space model. A novel kernel 
	kriged 
	Kalman filter  was developed using a deterministic RKHS approach. To 
	accommodate scenarios with limited prior 
	information,
	an 
	online multi-kernel learning technique was also developed to allow tracking of 
	the spatio-temporal 
	dynamics of the graph function. The structure of Laplacian kernels was
	exploited to achieve low computational complexity.
	\myitem\cmt{real test validation}Through numerical tests with synthetic as 
	well as real-data, the 
	novel algorithms were observed to perform markedly better than existing 
	alternatives. 
	\myitem\cmt{future work}Future work includes distributed implementations of the proposed filtering 
	algorithms, and data-driven 
	learning of 
	$\adjtransgraphmat\timetimenot{\timeind}{\timeind-1}$.
\end{myitemize}
%\appendix
%\subsection{Derivation of the kernel Kalman filter}
%\label{sec:derivation}
%\input{proofkrkf.tex}
%\small
\bibliographystyle{IEEEtranS}
\bibliography{my_bibliography}

\end{document}